\pdfoutput=1
\documentclass[11pt]{article}
\usepackage[final]{acl}
\usepackage{times}
\usepackage{multirow}
\usepackage{amssymb}
\usepackage{textcomp}
\usepackage{arydshln}
\usepackage[nointegrals]{wasysym}
\usepackage{graphicx}
\usepackage{tabularx}
\usepackage{amsmath}
\usepackage{algorithm}
\usepackage{algpseudocode}
\usepackage{tcolorbox}
\tcbuselibrary{listings,breakable}
\usepackage{xspace}
\usepackage{booktabs} 
\usepackage{tcolorbox}
\usepackage[hypcap=false]{caption}
\usepackage{latexsym}
\usepackage[T1]{fontenc}
\usepackage[utf8]{inputenc}
\usepackage{microtype}
\usepackage{amsmath}
\usepackage{cleveref}
\usepackage{algorithm}
\usepackage{algpseudocode}
\usepackage{inconsolata}
\usepackage{graphicx}
\usepackage{xcolor}
\usepackage{makecell} 
\usepackage{array}
\usepackage{colortbl}
\usepackage{xcolor}
\usepackage{pifont}
\newcommand{\reason}{\textcolor{orange!80!black}{\ensuremath{\bullet}}}

\newcommand{\cmark}{\ding{51}}
\newcommand{\xmark}{\ding{55}}
\definecolor{pos}{RGB}{215,48,39}   
\definecolor{neg}{RGB}{69,117,180}  
\definecolor{udqbg}{RGB}{219,209,233}
\definecolor{advorange}{RGB}{217,95,2}
\newcommand{\good}[1]{\textcolor{pos}{#1}}

\newcommand{\affmark}[1]{\textsuperscript{\normalsize #1}}
\usepackage[table]{xcolor}
\usepackage{comment}

\definecolor{mygreen}{RGB}{0, 153, 0}
\newcommand{\adv}[1]{\textcolor{advorange}{\textbf{#1}}}
\usepackage[table]{xcolor}
\usepackage{tikz}
\usepackage{tcolorbox}
\usetikzlibrary{tikzmark,calc}
\usepackage{enumitem}

\newcommand{\moveup}{\vspace*{-2mm}}
\newcommand{\moveups}{\vspace*{-1mm}}
\newcommand{\ignore}[1]{}
\newcommand{\pmark}{\raisebox{.1ex}{\small$\triangle$}}

\usepackage[utf8]{inputenc}
\usepackage{hyphenat}
\usepackage{xspace}
\usepackage{amsmath}
\usepackage{amsfonts}
\usepackage{hyperref}
\usepackage{url}
\usepackage{booktabs}
\usepackage{multirow}
\usepackage{makecell}
\usepackage{minibox}
\usepackage{bbm}
\usepackage{graphicx}
\usepackage{balance}
\usepackage{mathtools}
\usepackage{color}
\usepackage{marvosym}
\usepackage{ifthen}
\usepackage{textcomp}
\usepackage{enumitem}
\usepackage{verbatim}
\usepackage{algorithm}
\usepackage{numprint}
\usepackage{balance}

\newcommand{\chatoDisplayMode}[1]{#1}

\definecolor{MyRed}{rgb}{0.6,0.0,0.0} 
\definecolor{MyBlack}{rgb}{0.1,0.1,0.1} 
\newcommand{\inred}[1]{{\color{MyRed}\sf\textbf{\textsc{#1}}}}
\newcommand{\frameit}[2]{
  \begin{center}
  {\color{MyRed}
  \framebox[.9\columnwidth][l]{
    \begin{minipage}{.85\columnwidth}
    \inred{#1}: {\sf\color{MyBlack}#2}
    \end{minipage}
  }\\
  }
  \end{center}
}

\newcommand{\note}[2][]{\chatoDisplayMode{\def\@tmpsig{#1}\frameit{{\Pointinghand} Note}{#2\ifx \@tmpsig \@empty \else \mbox{ --\em #1}\fi}}}
\newcommand{\xhdr}[1]{\vspace{1.7mm}\noindent{{\bf #1.}}}

\newcommand{\textcite}[1]{\citeauthor{#1} \shortcite{#1}}

\newcommand{\dataset}{\textsc{TrekBench-Ctrl}\xspace}
\newcommand{\wild}{\textsc{TrekBench-Wild}\xspace}
\newcommand{\project}{\textsc{Trek}\xspace}
\newcommand{\tool}{\textsc{TrekTrace}\xspace}
\usepackage{fvextra}
\DefineVerbatimEnvironment{FrameworkPrompt}{Verbatim}{
  fontsize=\footnotesize,
  breaklines=true,
  breakanywhere=true,
  tabsize=2
}
\title{Where and Why Search-Augmented LLMs Fail When Reasoning over Rapidly Evolving Knowledge}

\author{
  Tianyi Hu\affmark{\quarternote} \quad
  Bhavyajeet Singh\affmark{\eighthnote,\twonotes \ensuremath{\dagger}} \quad
  Niket Tandon\affmark{\eighthnote} \quad
  Akhil Arora\affmark{\quarternote} \\
  \affmark{\quarternote} Aarhus University \quad
  \affmark{\eighthnote} Microsoft Research \quad
  \affmark{\twonotes} Technische Universität Darmstadt \\
  \texttt{\{tenney.hu, akhil.arora\}@cs.au.dk}
}
\begin{document}
\maketitle

\begingroup
\renewcommand{\thefootnote}{\textdagger}
\footnotetext{Work done during pre-PhD internship at Microsoft.}
\endgroup

\begin{abstract}
 As the world changes, new information is published while older records remain online. This complicates reasoning for large language models (LLMs), which must retrieve evidence and resolve conflicts to determine which information remains valid. 
Existing evaluations treat information seeking as a black box, revealing little about where and why LLMs fail.
To address this gap, we introduce \textbf{\project}\footnote{ https://github.com/au-clan/TREKBench}, a suite of \emph{three} components: \textbf{\tool} is a plug-and-play framework that attributes failures to specific stages of retrieval and reasoning. We apply it to \textbf{\wild}, a meta-benchmark of real-world questions that require complex reasoning over rapidly evolving knowledge. Surprisingly, the main bottleneck is post-retrieval reasoning rather than retrieval: frontier models are often misled by conflicting evidence even after retrieving all the information needed to answer.
To study this bottleneck, we build \textbf{\dataset}, a controlled benchmark with distractors spanning temporality, relevance, and trustworthiness, and show that benchmarks dominated by outdated distractors substantially underestimate task difficulty. Together, \project provides a testbed for diagnosing and improving reasoning over rapidly evolving knowledge.
\end{abstract}
\section{Introduction}

Enabling large language models (LLMs) to acquire and reason over evolving knowledge about the world and their users is a fundamental challenge in real-world deployment~\citep{huang2025survey}. The world and users change continuously~\citep{zhu2024freshbench}, whereas an LLM's parametric knowledge reflects an earlier snapshot~\citep{lazaridou2021mind}. This mismatch spans changes in \emph{world knowledge}~\citep{vu2024freshllms}, \emph{user-specific memories}~\citep{chao2026stale,uddin2026recall}, and \emph{external environments and tools}~\citep{chen2025learning}.

Across these settings, models often struggle to select temporally valid knowledge amid conflicting information~\citep{wu2024time}. Although temporal reasoning~\citep{uddin2025unseentimeqa} and knowledge conflicts~\citep{cattan2025dragged} have both been widely studied, they have largely been examined separately. However, rapidly evolving knowledge tightly couples them: outdated and updated versions of the same fact coexist, requiring models to reason about how facts evolve. Such reasoning is central to many everyday information needs, including news updates~\citep{liska2022streamingqa}, sports results~\citep{kasai2023realtime}, market information~\citep{yang2023fingpt}, weather forecasts~\citep{rasp2024weatherbench}, and crisis information~\citep{mccreadie2019trec}.

Retrieval-augmented generation (RAG)~\citep{lewis2020retrieval} works well for time-sensitive factual queries when the answer can be read directly from relevant, up-to-date evidence~\citep{chen2025question,lau2025reading}. The greater challenge is reasoning over what is retrieved: search results often mix conflicting versions of the same fact from different times with irrelevant or unreliable information (\emph{Figure~\ref{fig:main}(a)}), and RAG systems struggle to resolve these conflicts and select evidence valid at the target time~\citep{pham2025sealqa}.
Lacking scalable diagnostic tools, existing evaluations treat these failures as a black box:  they show \emph{that} a model fails, but not \emph{where} or \emph{why} the failure arises. This motivates three research questions:

\begin{figure*}[ht!]
\centering
\includegraphics[
  width=\linewidth,
]{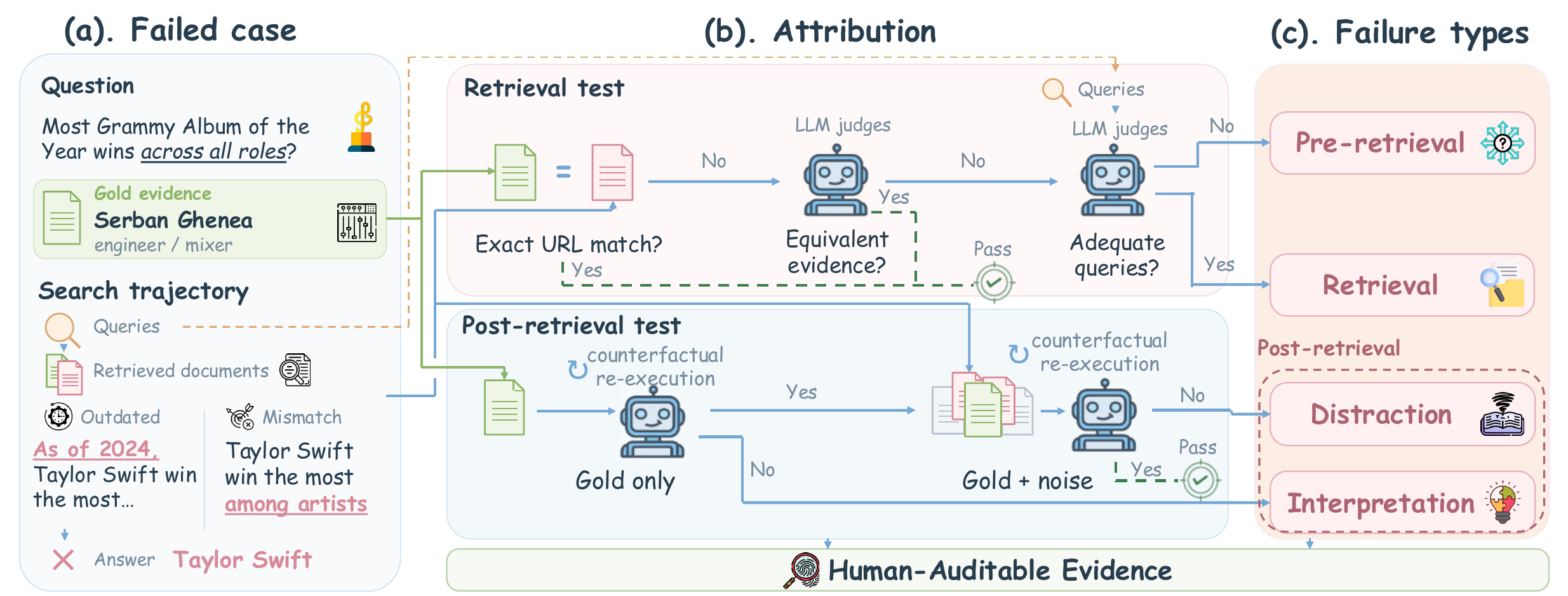}
\caption{Diagnosing failures in reasoning over evolving knowledge:
(a) Real-world failure case with two typical distractors: \emph{outdated information} and \emph{scope mismatch}.
(b) Overview of \textbf{\tool}, which automates attribution through retrieval analysis and counterfactual re-execution.
(c) Failure types with human-auditable evidence.}
\moveup
\moveup
\label{fig:main}
\end{figure*}

\begin{itemize}[
    leftmargin=1em,
    labelsep=0.2em,
    nosep
]
\item \textbf{RQ1 (Failure Attribution):} Can failures on rapidly evolving knowledge be attributed to stages of real-world search pipelines automatically and reliably, across models and datasets?
\item \textbf{RQ2 (Pipeline Bottlenecks):} Which pipeline stages contribute most to failures of frontier search-augmented models?
\item \textbf{RQ3 (Distraction Analysis):}  Which types of contextual information can mislead models toward wrong answers, and what factors shape their ability to resist such distraction?
\end{itemize}

To answer these questions, we introduce \textbf{\project} (\textbf{T}emporal \textbf{R}easoning over \textbf{E}volving \textbf{K}nowledge), consisting of three novel components:

\textbf{\tool} (\textbf{RQ1}) is an automated framework for diagnosing failures (Figure~\ref{fig:main}(b,c)).  It analyzes the search trajectory to check whether the answer-bearing evidence was retrieved, then reruns the model with that evidence to see whether it can answer correctly, attributing failure to missing evidence, incorrect reasoning, or both. Requiring no fine-grained annotations or retraining, the framework is plug-and-play across information-seeking tasks. Its attributions are supported by verifiable evidence and closely match human judgments.

To identify pipeline bottlenecks (\textbf{RQ2}), we construct \textbf{\wild}, a meta-benchmark for reasoning over rapidly evolving knowledge in real-world information seeking.  Surprisingly, contrary to prior findings that reasoning models remain robust to distracting evidence~\citep{zhang2026failure}, we find that post-retrieval reasoning is the dominant bottleneck when knowledge evolves rapidly: even frontier models are often misled by conflicting information despite having answer-bearing evidence.

We introduce \textbf{\dataset} (\textbf{RQ3}), a reproducible benchmark for investigating which distractors cause models to fail despite access to valid evidence. Its fixed contexts contain diverse distractors spanning temporality, relevance, and trustworthiness, testing conflict resolution beyond simply selecting the latest information. Our results show that benchmarks dominated by outdated distractors substantially underestimate task difficulty, while \dataset exposes persistent vulnerabilities in reasoning over rapidly evolving knowledge.

\section{Reasoning Over Rapidly Evolving Knowledge}
\label{sec:task}
In this section, we formalize the task objective and pipeline and identify its main failure modes.

\xhdr{Background and Task Formulation}
Reasoning over rapidly evolving knowledge involves two demanding objectives. 
First, the required facts often postdate the model's training cutoff, demanding \textit{access to up-to-date external information}~\cite{pham2025sealqa}. Second, answering requires \textit{strong reasoning capabilities} to disambiguate similar entities and integrate evidence across web pages. 
A common approach to addressing both requirements is
\textit{agentic search}, which interleaves reasoning with
iterative retrieval to gather relevant evidence and reason over it~\cite{xi2025infodeepseek,jin2025search}.

Formally, given a time-sensitive question $q$ with temporal
constraint $\tau_q$ (e.g., ``the latest''), model $\mathcal{M}$
iteratively reasons over accumulated evidence and issues
search queries $s_i$, receiving documents $\mathcal{D}_i=\mathcal{R}(s_i)$ from retriever $\mathcal{R}$, The process ends when $\mathcal{M}$ deems the evidence sufficient
or exhausts its budget. After $T$ rounds, the accumulated
evidence is $\mathcal{D}=\bigcup_{i=1}^{T}\mathcal{D}_i$,
where each document $d$ has a timestamp $t(d)$.
The model then selects relevant, temporally appropriate,
and reliable evidence from $\mathcal{D}$ to derive an answer
$a$ consistent with $\tau_q$.

\xhdr{Challenges and Failure Modes}
Agentic search combines external knowledge with reasoning. But it remains fragile: an error at any stage can lead to a wrong final answer. To analyze different failures, we summarize common \emph{system-induced} failure modes, as shown in Figure \ref{fig:main} (c):

\begin{itemize}[wide=3pt,nosep]
    \item \textbf{Pre-retrieval:} The model formulates a flawed query (e.g., with an incorrect temporal qualifier), so retrieval is misdirected.
    
      \item \textbf{Retrieval:} Despite a reasonable query, no answer-bearing document is returned (e.g., fresh pages matching $\tau_q$ are outranked by outdated ones).

    \item \textbf{Post-retrieval:} The answer-bearing evidence is retrieved, yet the model still fails, either through (i)~\emph{distraction}, where misleading context steers the answer away, or (ii)~\emph{interpretation}, where the model fails even on the gold evidence alone (e.g., misreading tables in the retrieved pages), revealing difficulty in reasoning over the evidence itself.
\end{itemize}

Beyond these \emph{system-induced} failures, \emph{system-independent} failures lie outside the system's control: the recorded judgment is wrong even though no pipeline stage is at fault (e.g., the ground-truth answer has shifted since annotation). We report such cases separately rather than attributing them to any pipeline stage.

\begin{table*}[!t]
\centering
\footnotesize
\setlength{\tabcolsep}{3pt}
\renewcommand{\arraystretch}{1.08}
\resizebox{\textwidth}{!}{%
\begin{tabular}{@{}l*{5}{c}@{\hspace{7pt}}*{5}{c}@{\hspace{7pt}}>{\columncolor{blue!7}}c>{\columncolor{blue!7}}c@{}}
\toprule
& \multicolumn{5}{c}{\cellcolor{yellow!25}Closed-book}
& \multicolumn{5}{c}{\cellcolor{pink!25}With Search}
& \multicolumn{2}{c}{\cellcolor{blue!18}Search Gain (pp) $\uparrow$} \\
\cmidrule(lr){2-6}\cmidrule(lr){7-11}\cmidrule(l){12-13}
Model
& FreshQA & InfoDS & SealQA & Micro & Macro
& FreshQA & InfoDS & SealQA & Micro & Macro
& $\Delta$Micro & $\Delta$Macro \\
\midrule
\textsc{Qwen3-8B}
& 26.5 & 3.2 & 4.7 & 11.8 & 11.5
& 34.7 & 0.0 & 9.4 & 16.0 & 14.7
& +4.2 & +3.2 \\

\textsc{Qwen3.8-27B}
& 44.9 & 0.0 & 3.1 & 16.7 & 16.0
& 71.4 & 22.6 & 25.0 & 40.3 & 39.7
& +23.6 & +23.7 \\

\textsc{GLM-4.7-Flash-30B}
& 36.7 & 0.0 & 1.6 & 13.2 & 12.8
& 55.1 & 9.7 & 12.5 & 26.4 & 25.8
& +13.2 & +13.0 \\

\textsc{Gemma-4-31B}
& 28.6 & 3.2 & 10.9 & 15.3 & 14.2
& 34.7 & 3.2 & 10.9 & 17.4 & 16.3
& +2.1 & +2.0 \\

\textsc{Qwen3-32B}
& 34.7 & 3.2 & 7.8 & 16.0 & 15.2
& 61.2 & 9.7 & 14.1 & 29.2 & 28.3
& +13.2 & +13.1 \\

\textsc{GPT-oss-120B}
& 32.7 & 3.2 & 4.7 & 13.9 & 13.5
& 69.4 & 16.1 & 17.2 & 34.7 & 34.2
& +20.8 & +20.7 \\

\textsc{Qwen3-235B}
& 40.8 & 3.2 & 9.4 & 18.8 & 17.8
& 63.3 & 9.7 & 17.2 & 31.2 & 30.0
& +12.5 & +12.2 \\
\midrule
\textsc{Claude-Sonnet-4.6}
& 49.0 & 12.9 & 12.5 & 25.0 & 24.8
& 75.5 & 45.2 & 35.9 & 51.4 & 52.2
& +26.4 & +27.4 \\

\textsc{Claude-Opus-4.6}
& 42.9 & 22.6 & 15.6 & 26.4 & 27.0
& 79.6 & 35.5 & 32.8 & 49.3 & 49.3
& +22.9 & +22.3 \\

\textsc{Gemini-3.1-Flash}
& 49.0 & 25.8 & 15.6 & 29.2 & 30.1
& 65.3 & 58.1 & 35.9 & 50.7 & 53.1
& +21.5 & +23.0 \\

\textsc{Gemini-3.1-Pro}
& 57.1 & 35.5 & 15.6 & 34.0 & 36.1
& 75.5 & 64.5 & 56.2 & 64.6 & 65.4
& +30.6 & +29.3 \\

\textsc{GPT-5.4-mini}
& 61.2 & 29.0 & 18.8 & 35.4 & 36.3
& 73.5 & 64.5 & 43.8 & 58.3 & 60.6
& +22.9 & +24.2 \\

\textsc{GPT-5.4}
& 69.4 & 54.8 & 29.7 & 48.6 & 51.3
& 85.7 & 77.4 & 54.7 & 70.1 & 72.6
& +21.5 & +21.3 \\
\bottomrule
\end{tabular}%
}
\caption{\emph{Accuracy} on \wild with high reasoning effort and w/. and w/o. web search.
Web search improves overall accuracy across models, but \wild remains challenging even for the strongest systems. Open-weight models still lag substantially behind proprietary models.
Full results refer to Table~\ref{tab:models-benchmarks-full}.}
\label{tab:models-benchmarks}
\vspace{-1.5em}
\end{table*}

\section{\wild: Where Do Models Fail?}
\label{sec:attr}

Search-augmented models still struggle to reason over rapidly evolving knowledge, yet the sources of their failures remain poorly understood. To investigate, we introduce \tool, an automated failure-attribution framework (\S\ref{sec:TemporalTrace}), and construct \wild, a curated meta-benchmark for reasoning over rapidly evolving knowledge (\S\ref{sec:Exp1}). We then apply \tool to \wild to identify where and why SoTA search-augmented models fail (\S\ref{sec:Results1}).

\subsection{\tool: A Failure Attribution Framework}

\label{sec:TemporalTrace}

\xhdr{Challenge}
Rapidly changing evidence complicates failure attribution: an incorrect answer may reflect missing evidence, faulty reasoning, or both, making it difficult to identify the true bottleneck.
Existing attribution methods are not directly suited to this setting: prior work~\citep{tian2026ragexplorer} require sentence-level gold-evidence annotations that are unavailable in most open-world information-seeking datasets, limiting their scalability, while others~\citep {jiao2026doctor} depend on training a dedicated diagnosis model and localize only the earliest faulty step, limiting adaptability and overlooking failures in later pipeline stages.

These limitations motivate three design requirements for automated failure attribution:
\textbf{R1: Adaptability}. The framework should transfer across datasets and pipelines without additional fine-grained evidence annotations, access to model internals, or task-specific retraining.
\textbf{R2: Comprehensive Coverage}. Retrieval and reasoning failures can coexist, so attribution should not stop at the first detected error.
\textbf{R3: Verifiability}. Each attribution should expose supporting evidence and diagnostic test outcomes for human verification.

\xhdr{Framework}
To meet these requirements, we introduce \textbf{\tool}, an automated framework that combines trajectory analysis with counterfactual reruns to diagnose failures in retrieving and reasoning over evidence. As illustrated in Figure~\ref{fig:main}(b), it uses the question, reference answer, gold evidence, and search trajectory without additional fine-grained annotations, access to model internals, or retraining, supporting adaptability across datasets and pipelines (\emph{R1}). It evaluates retrieval and post-retrieval behavior separately, allowing multiple attributions and diagnosing downstream reasoning failures even when retrieval fails, thereby ensuring coverage across pipeline stages (\emph{R2}). Specifically, it comprises two diagnostic tests:

\emph{(i) Retrieval test:}
This test determines whether the system obtained the evidence needed to answer and, if not, whether the gap stems from query formulation or retrieval.
We first check whether all gold-document URLs appear in the retrieved results. If so, the test passes. Otherwise, an LLM judge extracts the required answer-bearing information from unmatched gold documents and checks for equivalent evidence in the retrieved results. If all required information is covered, the test passes. If not, the judge assesses whether the issued queries adequately target the missing evidence: adequate queries indicate a \textbf{retrieval} failure, while inadequate queries indicate a \textbf{pre-retrieval} failure.

\emph{(ii) Post-retrieval test:}
This test distinguishes intrinsic reasoning limitations from failures induced by distracting context.
We first re-run answer generation with the evaluated model using only gold evidence.
If it fails, we label the failure \textbf{interpretation} and stop this test.
If it succeeds, we re-run with a \emph{gold-plus-noise} context: we retain the original retrieved context if it contains all gold or equivalent evidence; otherwise, we add the missing gold documents.
Failure in this rerun is labeled \textbf{distraction}, as the model succeeds with gold evidence alone but fails when additional documents are present. 

Alongside failure labels, \tool reports the supporting evidence from these tests, including matched documents, query assessments, and counterfactual outcomes, enabling the human verification requirement (\emph{R3}). Further implementation details of \tool are provided in Appendix~\ref{app:failure-framework}.

\subsection{Experimental Setup}
\label{sec:Exp1}
\xhdr{Benchmarks} 
No existing dataset focuses exclusively on reasoning over rapidly evolving knowledge. We address this gap with \textbf{\wild}, a meta-benchmark that aggregates data on rapidly evolving knowledge from three real-world information-seeking datasets. 
The three differ in reasoning difficulty: \texttt{FreshQA}~\citep{vu2024freshllms} contains general multi-hop questions;
\texttt{InfoDeepSeek}~\citep{xi2025infodeepseek} offers challenging questions purpose-built for multi-step agentic information seeking; and \texttt{SealQA}~\citep{pham2025sealqa} comprises the most difficult, expert-crafted questions, each iteratively refined until frontier models consistently fail. Details are shown in \S~\ref{app:subset-selection}.
 
\xhdr{Evaluated Systems}
We evaluate information-seeking systems that combine SoTA frontier and open-weight LLMs available at the time of our experiments with built-in or third-party search tools\footnote{https://www.tavily.com/}.
Specifically, we include \emph{GPT-5.4}~\citep{openai2026gpt54}, \emph{Gemini-3.1-Pro}~\citep{google2026gemini31}, \emph{Claude-Opus-4.6}~\citep{anthropic2026opus46}, and open-weight models from the \emph{Qwen}~\cite{qwen38,qwen3technicalreport}, \emph{GPT}~\cite{openai2025gptoss120bgptoss20bmodel}, \emph{GLM}~\cite{5team2025glm45agenticreasoningcoding}, and \emph{Gemma}~\cite{gemmateam2026gemma4} families, ranging from 8B to 235B parameters. The full model list is provided in \S~\ref{app: model_list} and settings in \S~\ref{app: details}.

\xhdr{Metric and Setups} Following prior work~\citep{pham2025sealqa,vu2024freshllms}, we report the \emph{accuracy} using each dataset's judging protocol. We additionally report micro-averaged accuracy over all examples and macro-averaged accuracy that weights each dataset equally.
For fine-grained error analysis, as described in \S~\ref{sec:TemporalTrace}, we first manually verify the reference answers and LLM-as-a-judge evaluations to distinguish system-independent from system-related errors. We then classify system-related errors according to the definitions in \S~\ref{sec:task}. 
Reproducibility details are shown in \S~\ref{app:wild_setup}.

\begin{figure*}[ht!]
\centering
\includegraphics[
  width=0.9\linewidth,
  trim={0mm 65mm 0mm 6mm},
  clip
]{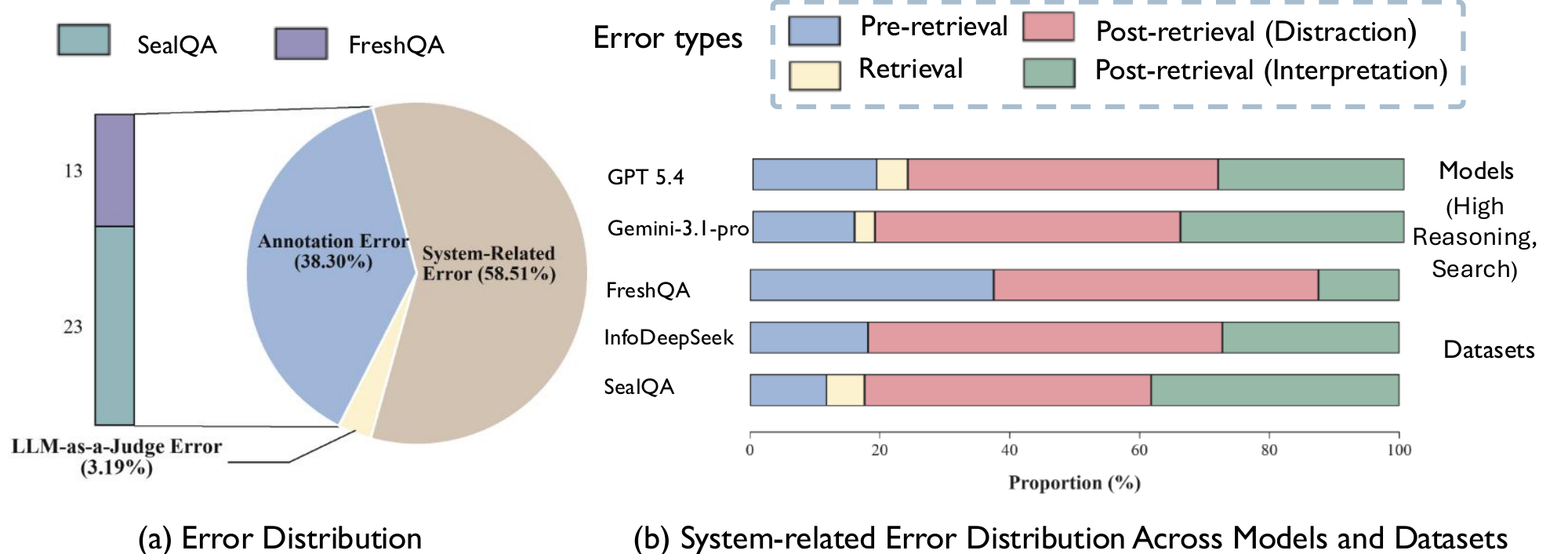}
\caption{\textbf{Overview of failure distributions} in \textbf{\wild} across high-reasoning, search-enabled models.}
\moveup
\label{fig:failure}
\moveup
\moveups
\end{figure*}

\subsection{Results and Analysis}
\label{sec:Results1}

\xhdr{Results}
Even leading frontier models with high reasoning effort and web search struggle to reason over rapidly evolving knowledge
(\Cref{tab:models-benchmarks}). \emph{GPT-5.4} achieves the highest micro-averaged accuracy (\emph{70.1\%}),
while \emph{Gemini-3.1-Pro} leads on SealQA, the hardest subset,
at only \emph{56.2\%}. Our experiments further show that open-weight models substantially
lag behind proprietary frontier models; Web search improves overall accuracy across all models, with particularly large gains for frontier models and open-weight models with strong reasoning capabilities, such as \emph{Qwen3.8-27B}.

\begin{figure}
    \centering
    \includegraphics[
        width=\linewidth,
        trim={0mm 2mm 0mm 0mm},
        clip
    ]{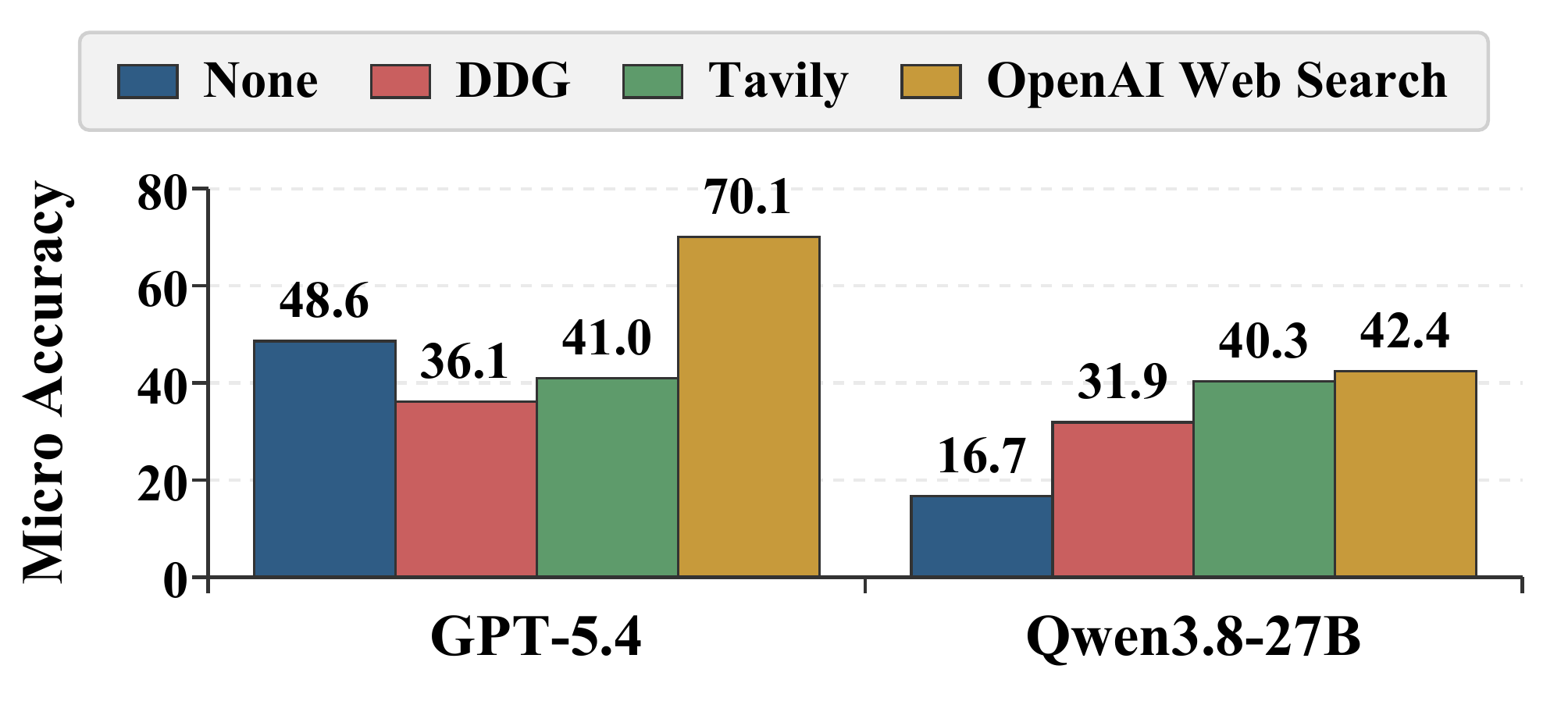}
    \caption{Micro Accuracy of \emph{GPT-5.4} and \emph{Qwen3.8-27B} under \emph{different search settings}, full results in Table~\ref{tab:search-comparison}.}
    \label{fig:search-comparison}
    \vspace{-1.5em}
\end{figure}

\xhdr{Analysis} 
\textbf{\tool} analysis of failures from
\emph{GPT-5.4-High-Search} and \emph{Gemini-3.1-Pro-High-Search},
the two leading models in Table~\ref{tab:models-benchmarks}    (Figure~\ref{fig:failure}), and comparisons across model--search combinations (Figure~\ref{fig:search-comparison}), yield \emph{four main findings}:

\begin{enumerate}[label=\textbf{\arabic*.},wide=0pt,nosep]
\item \textbf{\tool reliably diagnoses failures.}
The framework strongly agrees with human verification (Cohen's \(\kappa=0.930\), \S~\ref{sec:ver}). The few disagreements arise when \tool incorrectly flags a \emph{retrieval failure} even though other retrieved URLs already contain the same
information as the gold sources. This slightly overestimates
retrieval failures but does not change our main conclusions.

\item \textbf{Correct annotations can become outdated.}
Evolving knowledge accounts for most ($86\%$) of annotation errors: answers have changed ($76\%$), or gold pages no longer support the answers ($10\%$), highlighting the need for more thorough, ongoing verification of annotations~\cite{jiang2026benchmarks}.

\item \textbf{Better search alone is not enough.} As shown in \Cref{fig:search-comparison} and \Cref{tab:search-comparison},
the open-weight model achieves similar overall accuracy with Tavily and an OpenAI search proxy. In contrast, the frontier model GPT-5.4 performs substantially worse with third-party search tools than with its built-in search. These results highlight the importance of both strong reasoning models and effective search.

\item \textbf{Post-retrieval reasoning under distraction is the main bottleneck.}
Unlike prior work, which identifies retrieval as the main bottleneck~\cite{jiao2026doctor} or finds that reasoning models match gold-only performance even with noise~\cite{zhang2026failure}, we find that post-retrieval reasoning is the main bottleneck for frontier information-seeking systems when knowledge evolves rapidly.
This bottleneck is driven mainly by distraction from conflicting evidence, although interpretation errors become more prevalent as dataset difficulty increases. 
Of these two failure modes, interpretation errors reflect intrinsic reasoning limitations that have been widely studied~\cite{zhang2025survey,jin2025search}, whereas distraction from conflicting evidence remains underexplored.
This gap motivates our controlled benchmark, \dataset, for analyzing distractors in more depth (\S~\ref{sec:benchmark}).

\end{enumerate}

\begin{table*}[t]
\centering
\small
\setlength{\tabcolsep}{4pt}
\renewcommand{\arraystretch}{1}

\textbf{Query:} \emph{As of 2026, who has the most Grammy Album of the Year wins?}
\quad
\textbf{Answer:} \emph{Serban Ghenea} (mixer, five wins).
\par\vspace{3pt}

\begin{tabularx}{\textwidth}{@{}l>{\raggedright\arraybackslash}X@{}}
\toprule
\textbf{Distraction Type} & \textbf{Example} \\
\midrule

\texttt{Temporality}
& \good{[Before Feb.\ 2024]} \emph{Tom Coyne} and
\emph{Serban Ghenea} are tied with four wins. \\

\texttt{Relevance}
& [2026] \emph{Taylor Swift} has the most Album of the Year
wins \good{among artists}. \\

\texttt{Trustworthiness} (\emph{Republication})
& \good{[Unreliable source]} \adv{[Republished in 2026]}
\emph{Tom Coyne} and \emph{Serban Ghenea} are tied... \\

\texttt{Trustworthiness} (\emph{Fabrication})
& \good{[Unreliable source]} [2026]
\adv{\emph{John Hanes}} holds the record with six wins. \\

\bottomrule
\end{tabularx}
\moveup
\caption{Examples of distraction types.
\good{Red} marks distraction cues;
\adv{orange} marks adversarial edits.}
\label{tab:distractor-examples}
\moveup
\moveups
\end{table*}

\section{\dataset: Benchmarking under Controlled Distraction}
\label{sec:benchmark}

Post-retrieval distraction is a major failure source (\S~\ref{sec:attr}), but rapidly changing search results and web content make evaluations difficult to reproduce reliably.
Although existing benchmarks offer controlled settings, they rely almost exclusively on outdated distractors (\S~\ref{sec:rethink}). This narrow design rewards models for simply selecting the latest evidence and fails to capture the complexity of reasoning in real-world information seeking.
We therefore introduce \textbf{\dataset}, a long-context reasoning benchmark with fixed contexts and diverse, real-world-inspired distractors (\Cref{tab:distractor-examples} \& \S~\ref{sec:TemporalLens}). Finally, we evaluate reasoning models across distraction settings and test whether test-time mitigation improves robustness (\S~\ref{sec: res2}).

\subsection{Limitations of Existing Benchmarks}
\label{sec:rethink}
\begin{figure}
    \centering
    \includegraphics[width=1\linewidth]{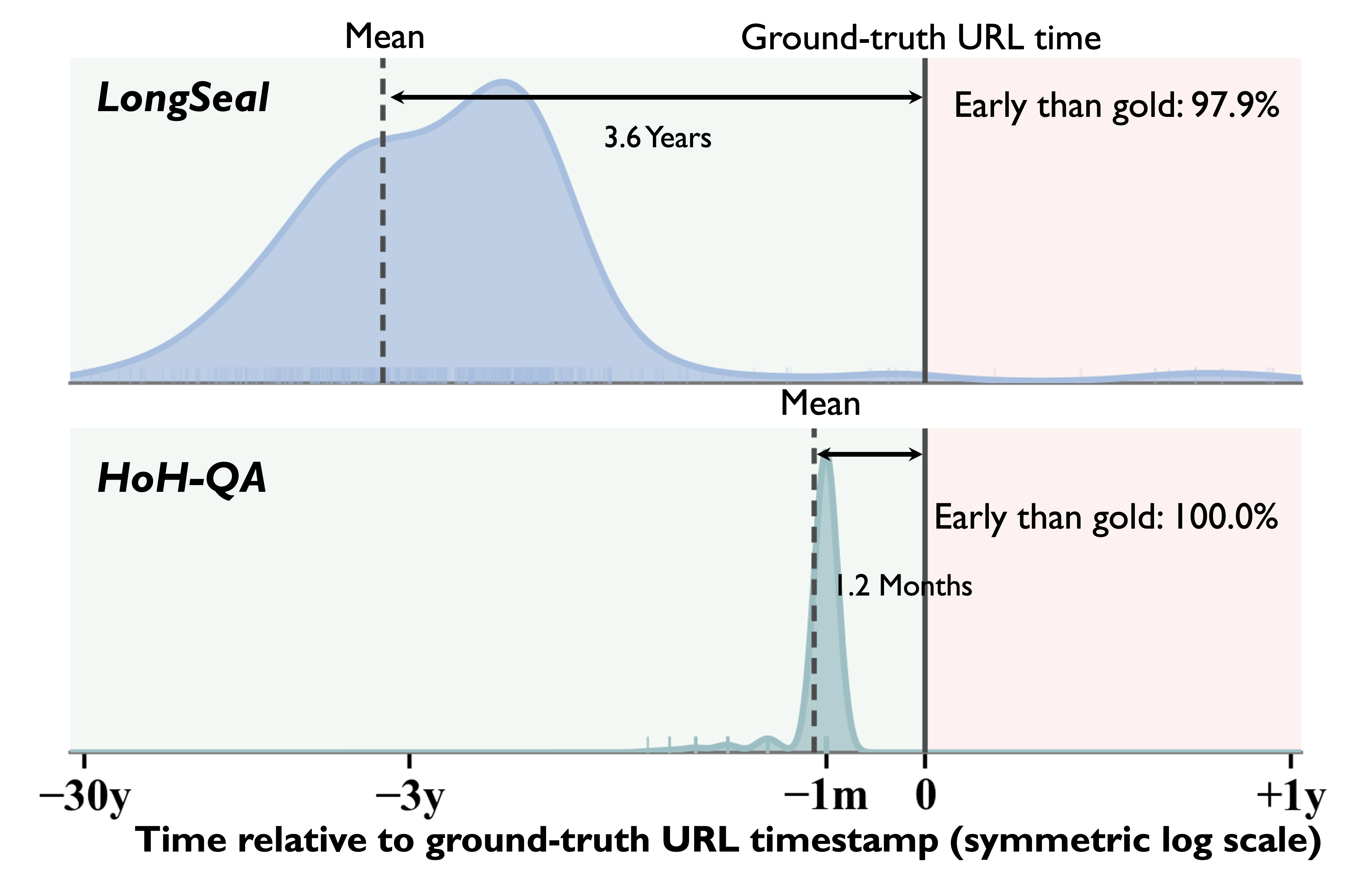}
    \caption{Pairwise time gaps reveal a weakness in existing benchmarks: nearly all distractors predate the ground-truth documents, so choosing the newest document is enough to answer correctly.}
    \label{fig:time_distribution}
    \moveup
    \moveup
    \moveups
\end{figure}

To our knowledge, \textit{LongSeal}~\cite{pham2025sealqa} and \textit{HoH-QA}~\cite{ouyang2025hoh} are the closest publicly available benchmarks to our setting: each includes a subset that evaluates reasoning over evolving knowledge using real-world documents with explicit timestamps.
Our analysis reveals that both rely almost exclusively on outdated distractors (Figure~\ref{fig:time_distribution}).
This imbalance limits challenge diversity and can
reward a simple shortcut: \emph{selecting the latest document
rather than resolving conflicting evidence}.
On the other hand, general distract information benchmarks~\citep{zhang2025magic}
do not capture the challenges specific to rapidly evolving knowledge.
E.g., \emph{NoisyBench}~\cite{lee2026lost} perturbations reduce SealQA
performance by only $\sim$1\%.
Together, these limitations (summarized in Table~\ref{tab:benchmark_comparison}) motivate a benchmark with more
diverse and challenging distractors for reasoning over evolving knowledge.

\subsection{Benchmark Construction}
\label{sec:TemporalLens}

To address the limitations in \S~\ref{sec:rethink}, we construct \textbf{\dataset}. 
Building on prior research on hard negatives in information-seeking systems~\cite{peikos2024systematic}, we target three key dimensions of distraction: \emph{Temporality}, \emph{Relevance}, and
\emph{Trustworthiness}, as illustrated in Table~\ref{tab:distractor-examples}.

 \textbf{Temporality}: Information was once correct but is outdated at the question's target time. As knowledge evolves, much information becomes outdated but remains accessible online. Models must reason across evidence to identify and discard outdated claims, even when
temporal cues are implicit (e.g., \textit{latest}). We retrieve real webpages that predate the gold documents and select those supporting answers different from the ground truth as distractors.

\textbf{Relevance}: Topically similar documents can provide plausible answers that do not satisfy the question's specific requirements. Especially when these documents are newer, models may mistake them for updates and incorrectly dismiss valid evidence as outdated.
To construct such distractors, we use an LLM to explore query variants by modifying constraints (e.g., replacing \emph{worldwide} with \emph{in the United States}  and retrieve real webpages supporting answers different from the ground truth. 
We favor pages newer than the gold one, making recency alone insufficient for selecting valid evidence.

\textbf{Trustworthiness}: Not all information is reliable. Beyond naturally occurring distractors, adversarial misinformation poses a further challenge to models' robustness to distraction.
Inspired by misinformation research, we construct two representative forms:
For \emph{Republication}, we refresh the publication dates of outdated documents without updating their claims, simulating old news republished to attract clicks~\cite{musi2022staying}.
For \emph{Fabrication}, we use an LLM to minimally rewrite gold evidence through entity substitutions, introducing plausible but false claims~\cite{jawahar2022automatic}.
Both forms introduce conflicting information, requiring models to reason across documents to identify reliable evidence.
We evaluate two settings: \emph{with domain cues}, following prior work~\cite{ahdritz2025smel}, synthetic documents carry suspicious fictitious domains (e.g., \texttt{totally-real-news.info}) that signal potential unreliability; and \emph{without domain cues}, where these explicit signals are hidden, and models must cross-check claims against other evidence in the context.

Detailed dataset construction is shown in \S~\ref{app:benchmark-construction}.

\providecommand{\dataset}{\textsc{TREKBench-CTRL}}
\providecommand{\dlt}[1]{{\scriptsize\textcolor{black!55}{$#1$}}}
\providecommand{\modelrow}[2]{%
  \rowcolor{black!6}
  \multicolumn{5}{@{\hspace{3pt}}l}{\textbf{#1}} &
  \multicolumn{4}{r@{\hspace{3pt}}}{\textit{Gold-Only:} #2} \\
}

\begin{table}[t]
\centering
\footnotesize
\setlength{\tabcolsep}{2pt}
\renewcommand{\arraystretch}{1.08}
\begin{tabularx}{\columnwidth}{@{}X *{4}{r@{\hspace{1.5pt}}l}@{}}
\toprule
\textbf{Method}
& \multicolumn{2}{c}{\textbf{OO}} & \multicolumn{2}{c}{\textbf{MND}}
& \multicolumn{2}{c}{\textbf{SAA}} & \multicolumn{2}{c@{}}{\textbf{SBA}} \\
\midrule
\modelrow{\textsc{GPT-5.4-mini} (High)}{70.8}
Direct     & 64.6 & \dlt{-6.3}  & 55.6 & \dlt{-15.3} & 43.1 & \dlt{-27.8} & 34.0 & \dlt{-36.8} \\
Rerank     & 62.5 & \dlt{-8.3}  & 54.9 & \dlt{-16.0} & 40.3 & \dlt{-30.6} & 33.3 & \dlt{-37.5} \\
Filter     & 62.5 & \dlt{-8.3}  & 52.8 & \dlt{-18.1} & 57.6 & \dlt{-13.2} & 43.1 & \dlt{-27.8} \\
Structured & 62.5 & \dlt{-8.3}  & 56.9 & \dlt{-13.9} & 35.4 & \dlt{-35.4} & 34.7 & \dlt{-36.1} \\
\midrule
\modelrow{\textsc{Qwen3.8-27B} (Thinking)}{70.1}
Direct     & 63.9 & \dlt{-6.3}  & 61.8 & \dlt{-8.3}  & 43.1 & \dlt{-27.1} & 34.7 & \dlt{-35.4} \\
Rerank     & 63.2 & \dlt{-6.9}  & 56.3 & \dlt{-13.9} & 36.8 & \dlt{-33.3} & 37.5 & \dlt{-32.6} \\
Filter     & 64.6 & \dlt{-5.6}  & 58.3 & \dlt{-11.8} & 59.0 & \dlt{-11.1} & 43.1 & \dlt{-27.1} \\
Structured & 58.3 & \dlt{-11.8} & 60.4 & \dlt{-9.7}  & 42.4 & \dlt{-27.8} & 35.4 & \dlt{-34.7} \\
\midrule
\modelrow{\textsc{GLM-4.7-Flash} (Thinking)$^{\dagger}$}{64.2}
Direct     & 56.8 & \dlt{-7.5}  & 51.4 & \dlt{-12.8} & 34.8 & \dlt{-29.5} & 28.6 & \dlt{-35.7} \\
Rerank     & 54.9 & \dlt{-9.3}  & 53.2 & \dlt{-11.0} & 30.8 & \dlt{-33.5} & 33.0 & \dlt{-31.2} \\
Filter     & 61.1 & \dlt{-3.2}  & 55.9 & \dlt{-8.4}  & 19.8 & \dlt{-44.4} & 18.9 & \dlt{-45.3} \\
Structured & 51.1 & \dlt{-13.1} & 44.1 & \dlt{-20.2} & 31.2 & \dlt{-33.0} & 21.7 & \dlt{-42.5} \\
\bottomrule
\end{tabularx}
\caption{Micro accuracy on \dataset\ under four distraction settings: Outdated Only (OO), Mixed Natural Distractors (MND), Source-Aware Adversarial (SAA), and Source-Blind Adversarial (SBA). Gray numbers are pp. changes from Gold-Only.
\emph{Direct}, \emph{Rerank}, \emph{Filter}, and \emph{Structured} denote Direct Generation and \emph{Reranking-Based}, \emph{Filtering-Based}, and \emph{Structured-Evidence} RAG Baselines. Outdated distractors alone underestimate the difficulty of reasoning over evolving knowledge, and no method closes the gap to Gold-Only on \dataset. Full results in \S~\ref{app:ctrl}.}
\label{tab:benchmark-main}
\vspace{-1.5em}
\end{table}

\subsection{Experiment Setups}

\xhdr{Datasets and Settings}
We retain the questions and evaluation protocols from \S~\ref{sec:Exp1}
and construct a diverse, challenging long context for each question
from pre-retrieved documents, following \S~\ref{sec:TemporalLens}.
Specifically, we evaluate four configurations
designed to introduce progressively stronger distraction:
(1) \emph{Outdated Only (OO)} uses only outdated distractors,
as previous datasets~\cite{pham2025sealqa};
(2) \emph{Mixed Natural Distractors (MND)} combines outdated
and irrelevant documents;
(3) \emph{Source-Aware Adversarial (SAA)} combines (2)
with republication and fabrication distractors,
and keeps source URLs visible as credibility cues;
(4) \emph{Source-Blind Adversarial (SBA)} preserves (3)'s
documents but hides source URLs.
Across settings, we use the same target distractor budget and context lengths roughly comparable, primarily varying the context composition to introduce more diverse and challenging distractors.
See \S~\ref{app:datasets_setup} for details.

\xhdr{Models} 
We evaluate three models from different families, including both
open-weight and proprietary models: \texttt{GPT-5.4-mini}~\cite{openai2026gpt54}\footnote{Following prior work~\cite{pham2025sealqa}, we exclude larger models such as GPT-5.4 and
Claude Opus 4.6, as it can cost nearly \$900 per model.},
and \texttt{Qwen3.8-27B}~\cite{qwen38} and \texttt{GLM-4.7-Flash}~\cite{5team2025glm45agenticreasoningcoding}. All models use high reasoning effort without web retrieval. 

\xhdr{Baselines} 
We use \emph{Direct Generation}, which answers from all provided
documents, as the standard baseline~\cite{pham2025sealqa}.
We further adapt three representative RAG strategies from related
tasks as \emph{strong baselines}:
(1) \emph{Reranking-Based RAG} reranks the retrieved documents
using joint semantic and temporal scores, then generates an answer
from the top-ranked documents~\cite{zhang2025mrag,ouyang2025hoh};
(2) \emph{Filtering-Based RAG} uses a reasoning model to filter
retrieved documents by relevance, requirement satisfaction, and
reliability before passing the retained context to the answer
generator~\cite{liu2024ra,wang2023learning};
(3) \emph{Structured-Evidence RAG} extracts candidate answers,
supporting evidence, and available timestamps and sources from
each document, then compares these records to resolve conflicts
and generate an answer~\cite{ranaldi2025improving,chiang2024metadata}.
Appendix~\ref{app:rag-baselines} provides details.
For each model, we also include \emph{Gold Only}, which provides
only the gold evidence as a reference for its best achievable performance without distractors.

\subsection{Results}
\label{sec: res2}
\begin{figure}
    \centering
    \includegraphics[width=1\linewidth]{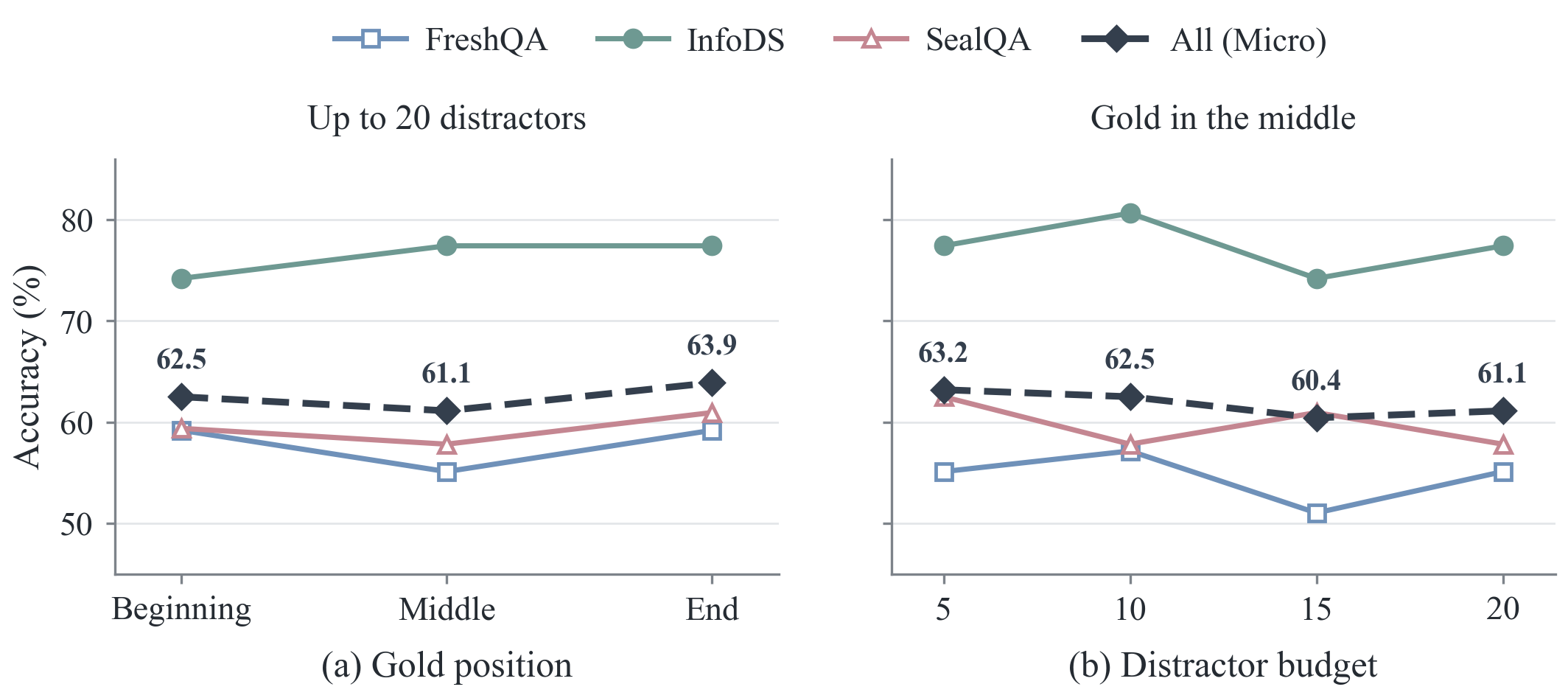}
    \caption{Effects of other factors on \dataset (GPT-5.4-mini under outdated only): (a) \emph{gold-document position} and (b) \emph{distractor budget}. Neither factor significantly affects accuracy, suggesting that distractor content, not number or position, drives the drops.}
    \label{fig:otherfactor}
    \vspace{-1em}
\end{figure}

We report results across models and
distraction settings in Table~\ref{tab:benchmark-main}.
Our main findings are as follows:

\begin{enumerate}[label=\textbf{\arabic*.},wide=2pt,nosep]
\item \textbf{Outdated distractors alone underestimate the difficulty of reasoning over evolving knowledge.}
Under OO, which mirrors the distractor composition of existing controlled benchmarks, GPT-5.4-mini with direct generation loses only 6.3 pp relative to Gold-Only (Table~\ref{tab:benchmark-main}). 
With more diverse distractors, the drop grows to $15.3$ pp under MND (\textbf{2.4}$\times$), $27.8$ pp under SAA (\textbf{4.4}$\times$), and $36.8$ pp under SBA (\textbf{5.8}$\times$). 
With a similar number and length of distractors across different settings (\Cref{tab:construction-counts}), and further experiments show that neither the position of the gold document nor the number of outdated distractors significantly affects accuracy (~\Cref{fig:otherfactor}), the gap therefore reflects the distractor types rather than context size or position, suggesting that benchmarks with only outdated distractors substantially underestimate the task's difficulty.
\item \textbf{Reasoning models know more than they use when knowledge evolves.}
Reasoning helps models judge evidence. When asked to judge each document explicitly, filtering-based RAG selects evidence much better than score-based reranking, revealing both the limits of reranking's shallow temporal and relevance scores, which distractors can easily match, and the benefit of reasoning over document content. Without being asked, however, models make far less use of this reasoning ability. 
\item \textbf{Models still tend to judge evidence by surface signals rather than content.}
Models are relatively robust to outdated distractors but lose about 9 pp once source URLs are hidden (SAA vs.\ SBA), suggesting that they rely on metadata such as timestamps and sources. To test this directly, we swap document metadata while keeping the content unchanged: swapping timestamps under outdated distractors and source domains under adversarial ones with the ground-truth significantly lowers accuracy by 7.3 and 23.6 pp, respectively. Models thus rely heavily on timestamps and sources, rather than content, to decide which evidence to trust.
\end{enumerate}

\section{Related Work}

\xhdr{Reasoning over Evolving Knowledge}
Time-sensitive question answering~\citep{piryani2025s} covers questions whose answers depend on time, either through explicit (e.g., \emph{in 2026}) or implicit (e.g., \emph{the latest}) constraints~\cite{meem2024pat}. These questions are especially challenging when they require reasoning over rapidly evolving real-world knowledge~\citep{vu2024freshllms, pham2025sealqa, nakshatri2025facts,li2025language}. 
To obtain and use this evidence, models must decompose questions into search queries~\cite{chen2025question}, retrieve temporally appropriate information~\cite{wu2024time}, and interpret and integrate the evidence while resisting contextual distraction~\cite{ouyang2025hoh}. Errors can arise at any of these stages, yet existing benchmarks~\cite{pham2025sealqa,vu2024freshllms} primarily report end-to-end accuracy, providing limited insight into where failures occur. This gap motivates our in-depth error analysis and automated failure attribution across the retrieval-augmented pipeline.

\xhdr{RAG Failure Diagnosis}
RAG diagnosis seeks to explain how errors arise across retrieval and reasoning.
In open-world information-seeking scenarios, however, existing attribution methods face limitations in adaptability and coverage.
Some require sentence-level gold evidence annotations~\citep{tian2026ragexplorer}, while others rely on dedicated diagnosis-model training~\citep{jiao2026doctor}, complicating reuse across datasets and models. Moreover, attribution often stops at the earliest faulty step~\citep{jiao2026doctor}, missing reasoning errors that co-occur with retrieval failures. These limitations motivate \tool, which combines trajectory analysis with independent retrieval and counterfactual reasoning tests, enabling multiple failure attributions without additional fine-grained annotations or task-specific training.

\xhdr{Knowledge Conflict and Contextual Distraction}
Knowledge conflicts fall into three categories: context-memory, inter-context, and intra-memory conflict~\citep{xu2024knowledge}. In rapidly evolving domains, retrieved contexts naturally mix contradictory information from different points in time~\cite{wu2024time}, making inter-context conflict our primary concern. 
 Since real-world retrieval results are constantly changing and thus hard to reproduce, benchmarks~\citep{pham2025sealqa,ouyang2025hoh} evaluate LLMs' ability to reason over collected contexts while resisting misleading information. However, the distractors in these datasets consist almost exclusively of outdated information, allowing models to succeed via a shortcut: always favoring the latest evidence.
Other benchmarks study conflicts from misinformation and ambiguity~\citep{cattan2025dragged, wang2025retrieval} or broader noise taxonomies for general RAG~\citep{zhang2025magic, wu2025pandora, lee2026lost}, but they target general queries or synthetic claims rather than fast-changing facts; E.g., the distraction in NoisyBench~\citep{lee2026lost} only reduces accuracy on SealQA by $\sim1\%$.
These gaps motivate a benchmark that combines diverse and challenging distractors over real, rapidly evolving knowledge; see \S~\ref{app:more_related_work} for a detailed comparison.

\section{Conclusion}
Reasoning over rapidly evolving knowledge requires models to retrieve fresh evidence and resolve conflicts to identify what remains valid.
We show that post-retrieval reasoning, rather than retrieval, is now the main bottleneck for search-augmented systems, and that models handle outdated distractors relatively well but remain vulnerable to more diverse distractors.
We hope \project moves evaluation beyond black-box toward understanding where and why models fail, and supports systems that reason about relevance, timeliness, and reliability together.

\clearpage

\section*{Limitations}

\xhdr{Dataset scale}
\project contains 144 English questions drawn from three established datasets. Although the statistical tests in Appendix~\ref{app: sta_test} support the validity of our findings, our experiments already cost about \$700 in total, the limit of our budget, so we could not enlarge the dataset further. We leave larger-scale datasets to future work.

\xhdr{Temporal granularity}
Although our dataset covers queries from diverse domains, such as sports, finance, and politics, it does not include information that changes by the minute or second, such as stock prices; the facts it covers typically change over days or weeks. We believe that minute- or second-level information requires specialized tools to query rather than general web search, which is beyond our scope, and we leave it to future work.

\xhdr{Distractor construction}
Republication and fabrication distractors, in particular, are constructed programmatically or with LLMs and verified automatically, so they may not reflect the full variety of real-world misinformation. Two annotators additionally checked 30 random questions and found no case in which the distractors prevented them from identifying the correct answer. However, we cannot rule out that the generated distractors introduce additional ambiguity that makes the correct answer unidentifiable for some questions.

\xhdr{Mitigation strategies}
Our evaluation focuses on end-to-end LLM information-seeking systems and general test-time strategies. Given the limited scale of our current data, we leave training-based approaches to future work.

\section*{Acknowledgements}
We sincerely thank Tianhui (Stella) Xu for her insightful suggestions and careful proofreading. 
Arora’s lab is partly supported by grants from the Novo Nordisk Foundation (NNF24OC0099109), the Pioneer Centre for AI, and EU Horizon 2020 (101168951). 
We also gratefully acknowledge generous gifts from Microsoft and IT-vest - networking universities.

\bibliography{custom}

\appendix
\clearpage

\begin{table*}[!t]
\centering
\footnotesize

\setlength{\tabcolsep}{3pt}
\renewcommand{\arraystretch}{1.08}
\resizebox{\textwidth}{!}{%
\begin{tabular}{@{}l*{5}{c}@{\hspace{7pt}}*{5}{c}@{\hspace{7pt}}>{\columncolor{blue!7}}c>{\columncolor{blue!7}}c@{}}
\toprule
& \multicolumn{5}{c}{\cellcolor{yellow!25}\textbf{Closed-book}}
& \multicolumn{5}{c}{\cellcolor{pink!25}\textbf{With Search}}
& \multicolumn{2}{c}{\cellcolor{blue!18}\textbf{Search Gain (pp)} $\uparrow$} \\
\cmidrule(lr){2-6}\cmidrule(lr){7-11}\cmidrule(l){12-13}
\textbf{Model}
& \textbf{FreshQA} & \textbf{InfoDS} & \textbf{SealQA} & \textbf{Micro} & \textbf{Macro}
& \textbf{FreshQA} & \textbf{InfoDS} & \textbf{SealQA} & \textbf{Micro} & \textbf{Macro}
& $\Delta$\textbf{Micro} & $\Delta$\textbf{Macro} \\
\midrule
\textsc{Qwen3-8B} \reason
& 26.5 & 3.2 & 4.7 & 11.8 & 11.5
& 34.7 & 0.0 & 9.4 & 16.0 & 14.7
& +4.2 & +3.2 \\
\textsc{Gemma-4-31B} \reason
& 28.6 & 3.2 & 10.9 & 15.3 & 14.2
& 34.7 & 3.2 & 10.9 & 17.4 & 16.3
& +2.1 & +2.0 \\
\textsc{Qwen3-32B} \reason
& 34.7 & 3.2 & 7.8 & 16.0 & 15.2
& 61.2 & 9.7 & 14.1 & 29.2 & 28.3
& +13.2 & +13.1 \\
\textsc{Qwen3-235B} \reason
& 40.8 & 3.2 & 9.4 & 18.8 & 17.8
& 63.3 & 9.7 & 17.2 & 31.2 & 30.0
& +12.5 & +12.2 \\
\textsc{GPT-oss-120B} \reason
& 32.7 & 3.2 & 4.7 & 13.9 & 13.5
& 69.4 & 16.1 & 17.2 & 34.7 & 34.2
& +20.8 & +20.7 \\
\textsc{GLM-4.7-Flash} \reason
& 36.7 & 0.0 & 1.6 & 13.2 & 12.8
& 55.1 & 9.7 & 12.5 & 26.4 & 25.8
& +13.2 & +13.0 \\
\textsc{Qwen3.8-27B} \reason
& 44.9 & 0.0 & 3.1 & 16.7 & 16.0
& 71.4 & 22.6 & 25.0 & 40.3 & 39.7
& +23.6 & +23.7 \\

\midrule
\textsc{Claude-Sonnet-4.6}
& 49.0 & 19.4 & 9.4 & 25.0 & 25.9
& 67.3 & 51.6 & 21.9 & 43.8 & 46.9
& +18.8 & +21.0 \\
\textsc{Claude-Sonnet-4.6} \reason
& 49.0 & 12.9 & 12.5 & 25.0 & 24.8
& 75.5 & 45.2 & 35.9 & 51.4 & 52.2
& +26.4 & +27.4 \\
\textsc{Claude-Opus-4.6}
& 40.8 & 9.7 & 10.9 & 20.8 & 20.5
& 69.4 & 48.4 & 35.9 & 50.0 & 51.2
& +29.2 & +30.8 \\
\textsc{Claude-Opus-4.6} \reason
& 42.9 & 22.6 & 15.6 & 26.4 & 27.0
& \underline{79.6} & 35.5 & 32.8 & 49.3 & 49.3
& +22.9 & +22.3 \\

\addlinespace[3pt]
\textsc{Gemini-3.1-Flash}
& 38.8 & 12.9 & 6.2 & 18.8 & 19.3
& 69.4 & 45.2 & 23.4 & 43.8 & 46.0
& +25.0 & +26.7 \\
\textsc{Gemini-3.1-Flash} \reason
& 49.0 & 25.8 & 15.6 & 29.2 & 30.1
& 65.3 & 58.1 & 35.9 & 50.7 & 53.1
& +21.5 & +23.0 \\
\textsc{Gemini-3.1-Pro}
& 57.1 & 29.0 & 9.4 & 29.9 & 31.9
& 67.3 & \underline{64.5} & 35.9 & 52.8 & 55.9
& +22.9 & +24.1 \\
\textsc{Gemini-3.1-Pro} \reason
& 57.1 & \underline{35.5} & 15.6 & 34.0 & 36.1
& 75.5 & \underline{64.5} & \textbf{56.2} & \underline{64.6} & \underline{65.4} & +30.6 & +29.3 \\

\addlinespace[3pt]
\textsc{GPT-5.4-mini}
& 49.0 & 12.9 & 6.2 & 22.2 & 22.7
& 61.2 & 29.0 & 20.3 & 36.1 & 36.9
& +13.9 & +14.1 \\
\textsc{GPT-5.4-mini} \reason
& \underline{61.2} & 29.0 & \underline{18.8} & \underline{35.4} & \underline{36.3}
& 73.5 & \underline{64.5} & 43.8 & 58.3 & 60.6
& +22.9 & +24.2 \\
\textsc{GPT-5.4}
& 44.9 & 12.9 & 10.9 & 22.9 & 22.9
& 65.3 & 48.4 & 26.6 & 44.4 & 46.8
& +21.5 & +23.8 \\
\textsc{GPT-5.4} \reason
& \textbf{69.4} & \textbf{54.8} & \textbf{29.7} & \textbf{48.6} & \textbf{51.3}
& \textbf{85.7} & \textbf{77.4} & \underline{54.7} & \textbf{70.1} & \textbf{72.6}
& +21.5 & +21.3 \\
\bottomrule
\end{tabular}%
}
\caption{Accuracy across the three subsets of \wild, with and without
web search.  Micro averages over all examples; Macro averages equally across datasets. 
\reason~denotes high reasoning effort. \textbf{Bold} and \underline{underlined} indicate the best and
second-best results in each column.}
\label{tab:models-benchmarks-full}
\end{table*}

\begin{table}
\centering
\footnotesize
\setlength{\tabcolsep}{3pt}
\renewcommand{\arraystretch}{1.05}

\resizebox{\columnwidth}{!}{%
\begin{tabular}{@{}lccccc@{}}
\toprule
\textbf{Search} & \textbf{FreshQA} & \textbf{InfoDS}
& \textbf{SealQA} & \textbf{Micro} & \textbf{Macro} \\
\midrule
\multicolumn{6}{c}{\textsc{GPT-5.4}} \\
\midrule

None
& 69.4 & 54.8 & 29.7 & 48.6 & 51.3 \\
DDG
& 63.3 & 16.1 & 25.0 & 36.1 & 34.8 \\
Tavily 
& 67.3 & 19.4 & 31.2 & 41.0 & 39.3 \\
OpenAI Web Search
& \textbf{85.7} & \textbf{77.4} & \textbf{54.7}
& \textbf{70.1} & \textbf{72.6} \\
Claude Web Search
& 65.3 & 32.3 & 31.2 & 43.1 & 42.9 \\

\midrule
\multicolumn{6}{c}{\textsc{Qwen3.8-27B}} \\
\midrule

None
& 44.9 & \phantom{0}0.0 & \phantom{0}3.1 & 16.7 & 16.0 \\
DDG
& 65.3 & 12.9 & 15.6 & 31.9 & 31.3 \\
Tavily
& 71.4 & \textbf{22.6} & 25.0 & 40.3 & 39.7 \\
OpenAI Web Search
& 71.4 & 16.1 & 32.8 & 42.4 & 40.1 \\
Claude Web Search
& \textbf{77.6} & \textbf{22.6} & \textbf{35.9}
& \textbf{47.2} & \textbf{45.4} \\

\bottomrule
\end{tabular}%
}
\caption{Effect of search configuration on accuracy (\%) across
the three subsets of \wild.
Micro averages over examples; Macro averages equally across datasets.
DDG denotes DuckDuckGo.
\textbf{Bold} indicates the best result within each model.}
\label{tab:search-comparison}
\end{table}

\begin{table}
\centering
\footnotesize
\setlength{\tabcolsep}{5pt}

\begin{tabular}{@{}lrr@{}}
\toprule
\textbf{Setting} & \textbf{All (64)} & \textbf{Matched (61)} \\
\midrule
\multicolumn{3}{c}{\textit{Gold position: 20-distractor budget}} \\
\midrule
Beginning & 59.4 & 62.3 \\
Middle    & 57.8 & 60.7 \\
End       & 60.9 & 63.9 \\
\midrule
\multicolumn{3}{c}{\textit{Distractor count: gold in the middle}} \\
\midrule
5  & 62.5 & 63.9 \\
10 & 57.8 & 60.7 \\
15 & 60.9 & 63.9 \\
20 & 57.8 & 60.7 \\
\bottomrule
\end{tabular}
\caption{Accuracy (\%) under different gold positions and
distractor budgets. All includes all 64 questions; Matched
includes the same 61 questions with 20 available distractors.
Shortages are retained in All. The middle/20 condition is shared
between the two experiments.}
\label{tab:position-count}
\end{table}

\begin{figure}
    \centering
    \includegraphics[width=1\linewidth]{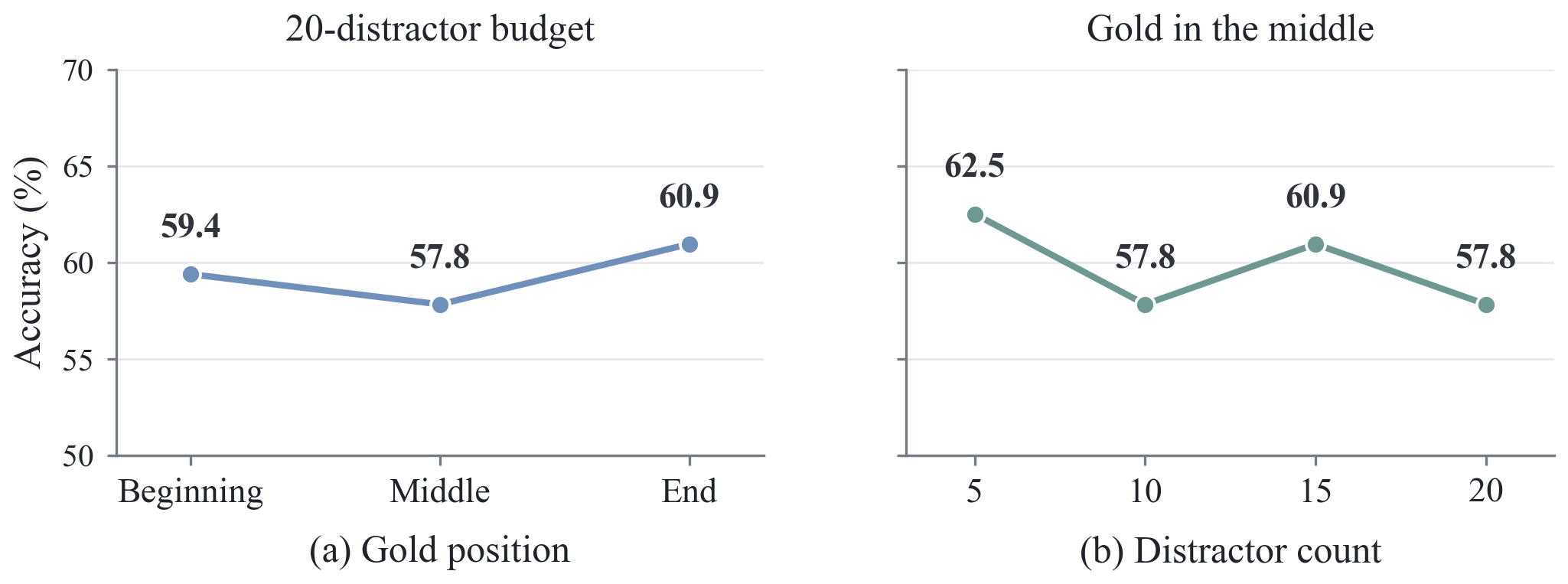}
    \caption{Effects of gold position and distractor budget on
GPT-5.4-mini-High. Accuracy does not differ significantly across gold positions or distractor counts.}
    \label{fig: position}
\end{figure}

\section{Additional Experimental Results}
\label{app: full_results}
\subsection{Additional Experimental Results on \wild}

\xhdr{Extended Model Comparison}
Table~\ref{tab:models-benchmarks-full} reports the complete results on \wild, including additional open-weight baselines and both micro- and macro-averaged accuracy. 

\xhdr{Effect of Search Configuration}
Table~\ref{tab:search-comparison} shows that enabling search does not necessarily improve performance. GPT-5.4 achieves 70.1\% micro-averaged accuracy with OpenAI Web Search, compared with 48.6\% without search, but drops to 36.1\% with DDG and 41.0\% with Tavily. Qwen3.8-27B benefits from all tested search configurations, reaching 42.4\% with OpenAI Web Search, 40.3\% with Tavily, and 31.9\% with DDG. These results highlight persistent differences between open-source and closed-source models in their use of search tools.

\subsection{Additional Experimental Results on \dataset}
\label{app:ctrl}
\providecommand{\dataset}{\textsc{TREKBench-CTRL}}
\colorlet{microbg}{blue!7}
\providecommand{\dlt}[1]{{\scriptsize\textcolor{black!55}{$#1$}}}
\providecommand{\modelrow}[2]{%
  \rowcolor{black!6}
  \multicolumn{13}{@{}l}{\textbf{#1}} &
  \multicolumn{8}{c@{}}{\textit{Gold-Only:} #2} \\
}

\providecommand{\tbdrow}[1]{%
  #1 & -- & -- & -- & -- & -- & -- & -- & -- & -- & -- & -- & --
  & \multicolumn{2}{>{\columncolor{microbg}}c}{--} & \multicolumn{2}{>{\columncolor{microbg}}c}{--}
  & \multicolumn{2}{>{\columncolor{microbg}}c}{--} & \multicolumn{2}{>{\columncolor{microbg}}c@{}}{--} \\
}

\begin{table*}[t]
\centering
\footnotesize
\setlength{\tabcolsep}{2.2pt}
\renewcommand{\arraystretch}{1.1}
\resizebox{\textwidth}{!}{%
\begin{tabular}{@{}l *{12}{c}
  *{3}{>{\columncolor{microbg}[\tabcolsep][0pt]}r@{}%
       >{\columncolor{microbg}[0pt][\tabcolsep]\hspace{1.5pt}}l<{\hspace{3pt}}}
  >{\columncolor{microbg}[\tabcolsep][0pt]}r@{}%
  >{\columncolor{microbg}[0pt][0pt]\hspace{1.5pt}}l@{}}
\toprule
& \multicolumn{4}{c}{\textbf{FreshQA}}
& \multicolumn{4}{c}{\textbf{InfoDS}}
& \multicolumn{4}{c}{\textbf{SealQA}}
& \multicolumn{8}{>{\columncolor{microbg}}c@{}}{\textbf{Micro Avg.}} \\
\cmidrule(lr){2-5}\cmidrule(lr){6-9}\cmidrule(lr){10-13}\cmidrule(l){14-21}
\textbf{Method}
& OO & MND & SAA & SBA
& OO & MND & SAA & SBA
& OO & MND & SAA & SBA
& \multicolumn{2}{>{\columncolor{microbg}}c}{OO} & \multicolumn{2}{>{\columncolor{microbg}}c}{MND}
& \multicolumn{2}{>{\columncolor{microbg}}c}{SAA} & \multicolumn{2}{>{\columncolor{microbg}}c@{}}{SBA} \\
\midrule
\modelrow{\textsc{GPT-5.4-mini} (High)}{70.8}
Direct Generation
& 61.2 & 55.1 & 40.8 & 38.8
& 80.6 & 74.2 & 54.8 & 51.6
& 59.4 & 46.9 & 39.1 & 21.9
& 64.6 & \dlt{-6.3} & 55.6 & \dlt{-15.3} & 43.1 & \dlt{-27.8} & 34.0 & \dlt{-36.8} \\
Reranking-Based RAG
& 57.1 & 57.1 & 42.9 & 40.8
& 77.4 & 71.0 & 64.5 & 51.6
& 59.4 & 45.3 & 26.6 & 18.8
& 62.5 & \dlt{-8.3} & 54.9 & \dlt{-16.0} & 40.3 & \dlt{-30.6} & 33.3 & \dlt{-37.5} \\
Filtering-Based RAG
& 55.1 & 53.1 & 53.1 & 42.9
& 77.4 & 64.5 & 74.2 & 61.3
& 60.9 & 46.9 & 53.1 & 34.4
& 62.5 & \dlt{-8.3} & 52.8 & \dlt{-18.1} & 57.6 & \dlt{-13.2} & 43.1 & \dlt{-27.8} \\
Structured-Evidence RAG
& 55.1 & 51.0 & 30.6 & 26.5
& 77.4 & 74.2 & 51.6 & 58.1
& 60.9 & 53.1 & 31.3 & 29.7
& 62.5 & \dlt{-8.3} & 56.9 & \dlt{-13.9} & 35.4 & \dlt{-35.4} & 34.7 & \dlt{-36.1} \\
\midrule
\modelrow{\textsc{Qwen3.8-27B} (Thinking)}{70.1}
Direct Generation
& 61.2 & 63.3 & 38.8 & 32.7
& 74.2 & 80.6 & 67.7 & 61.3
& 60.9 & 51.6 & 34.4 & 23.4
& 63.9 & \dlt{-6.3} & 61.8 & \dlt{-8.3} & 43.1 & \dlt{-27.1} & 34.7 & \dlt{-35.4} \\
Reranking-Based RAG
& 63.3 & 61.2 & 36.7 & 36.7
& 74.2 & 74.2 & 58.1 & 64.5
& 57.8 & 43.8 & 26.6 & 25.0
& 63.2 & \dlt{-6.9} & 56.3 & \dlt{-13.9} & 36.8 & \dlt{-33.3} & 37.5 & \dlt{-32.6} \\
Filtering-Based RAG
& 57.1 & 55.1 & 53.1 & 38.8
& 80.6 & 71.0 & 77.4 & 61.3
& 62.5 & 54.7 & 54.7 & 37.5
& 64.6 & \dlt{-5.6} & 58.3 & \dlt{-11.8} & 59.0 & \dlt{-11.1} & 43.1 & \dlt{-27.1} \\
Structured-Evidence RAG
& 49.0 & 57.1 & 38.8 & 28.6
& 71.0 & 77.4 & 58.1 & 54.8
& 59.4 & 54.7 & 37.5 & 31.3
& 58.3 & \dlt{-11.8} & 60.4 & \dlt{-9.7} & 42.4 & \dlt{-27.8} & 35.4 & \dlt{-34.7} \\
\midrule
\modelrow{\textsc{GLM-4.7-Flash} (Thinking)$^{\dagger}$}{64.2}
Direct Generation
& 47.9 & 43.2 & 31.0 & 26.7
& 75.0 & 66.7 & 57.7 & 46.4
& 56.5 & 51.3 & 16.7 & 15.6
& 56.8 & \dlt{-7.5} & 51.4 & \dlt{-12.8} & 34.8 & \dlt{-29.5} & 28.6 & \dlt{-35.7} \\
Reranking-Based RAG
& 50.0 & 47.7 & 19.5 & 26.7
& 64.0 & 70.8 & 57.7 & 51.7
& 54.9 & 48.8 & 20.8 & 24.1
& 54.9 & \dlt{-9.3} & 53.2 & \dlt{-11.0} & 30.8 & \dlt{-33.5} & 33.0 & \dlt{-31.2} \\
Filtering-Based RAG
& 53.5 & 43.6 & 28.9 & 29.3
& 76.2 & 75.0 & 63.2 & 40.9
& 61.3 & 70.6 & 1.6 & 4.7
& 61.1 & \dlt{-3.2} & 55.9 & \dlt{-8.4} & 41.4 & \dlt{-22.9} & 36.4 & \dlt{-27.9} \\
Structured-Evidence RAG
& 46.9 & 33.3 & 30.6 & 22.9
& 61.3 & 65.4 & 41.4 & 37.9
& 49.1 & 42.6 & 25.5 & 13.1
& 51.1 & \dlt{-13.1} & 44.1 & \dlt{-20.2} & 31.2 & \dlt{-33.0} & 21.7 & \dlt{-42.5} \\
\bottomrule
\end{tabular}%
}
\caption{Accuracy (\%) on \dataset\ under four distraction settings:
Outdated Only (OO), Mixed Natural Distractors (MND), Source-Aware Adversarial (SAA), and Source-Blind Adversarial (SBA).
Micro Avg.\ (shaded) averages over all questions across subsets; small gray numbers give percentage-point changes from Gold-Only.}
\label{tab:benchmark-ctrl-full}
\end{table*}

See ~\Cref{tab:benchmark-ctrl-full} for the full results.

\begin{table}[t]
\centering
\footnotesize

\setlength{\tabcolsep}{6pt}
\begin{tabular}{@{}lrrr@{}}
\toprule
\textbf{Year} & \textbf{FreshQA} & \textbf{InfoDS} & \textbf{SealQA} \\
\midrule
2022        & 3  & 0  & 0 \\
2023        & 6  & 0  & $2^{\dagger}$ \\
2024        & 7  & 0  & 8 \\
2025        & 33 & 25 & 33 \\
2026        & 0  & 0  & 21 \\
Unspecified & 0  & 6  & 0 \\
\midrule
\textbf{Total} & \textbf{49} & \textbf{31} & \textbf{64} \\
\bottomrule
\end{tabular}

\smallskip
\parbox{\columnwidth}{\scriptsize
$\dagger$ Originally labeled ``before 2024''; grouped under
2023 as an upper-bound approximation.
}
\caption{Distribution of temporal reference years across the three subsets.}
\label{tab:dataset-time}
\end{table}

\subsection{Effects of Gold Position and Distractor Count in \dataset}
\label{app: effect}

\paragraph{Gold position.}
As shown in \Cref{fig: position}, accuracy ranges from 57.8\% to 60.9\% across positions.
Paired exact McNemar tests find no significant differences
(all $p \geq 0.625$), providing no clear evidence of a systematic
position effect in this setting.

\paragraph{Distractor count.}
As shown in \Cref{fig: position}, accuracy does not decrease monotonically as the distractor
budget increases: it reaches 62.5\%, 57.8\%, 60.9\%, and 57.8\%
with 5, 10, 15, and 20 distractors, respectively. 
Cochran's Q test across the 64 paired questions detects no
significant difference among these settings
($Q(3)=2.19$, $p=0.534$).

\subsection{Statistical Significance and Confidence Interval Analysis}
\label{app: sta_test}
\subsubsection{Statistical Significance of Distractor Effects}
\label{app:ctrl-significance}

\paragraph{Protocol.}
We analyze for each condition $X \in \{\mathrm{NMD,SAA,SBA}\}$, we test
\[
H_0: \mathrm{Acc}_{OO}=\mathrm{Acc}_{X},
\qquad
H_1: \mathrm{Acc}_{OO}\neq\mathrm{Acc}_{X}.
\]
We use two-sided exact McNemar tests on paired correctness
outcomes, with Holm correction across the three comparisons
at $\alpha=0.05$.
Pointwise 95\% confidence intervals for accuracy reductions
are estimated using 200,000 paired bootstrap resamples.

\begin{table}[t]
\centering
\footnotesize
\setlength{\tabcolsep}{3pt}
\caption{Paired comparisons against OO.
Accuracy is reported in percent; $\Delta$ denotes the
decrease from OO in percentage points.
$p_{\mathrm{Holm}}$ denotes the adjusted $p$-value.}
\label{tab:ctrl-significance}
\begin{tabular}{@{}lrrcc@{}}
\toprule
Condition & Acc. & $\Delta$ & 95\% CI of $\Delta$
& $p_{\mathrm{Holm}}$ \\
\midrule
OO  & 64.6 & --   & --           & -- \\
NMD & 55.6 & 9.0  & [1.4, 16.7]  & $0.029$ \\
SAA & 43.1 & 21.5 & [13.9, 29.9] & $1.57\times10^{-6}$ \\
SBA & 34.0 & 30.6 & [22.2, 38.9] & $3.92\times10^{-10}$ \\
\bottomrule
\end{tabular}
\end{table}

\paragraph{Results.}
As shown in Table~\ref{tab:ctrl-significance}, all three
conditions significantly reduce accuracy relative to OO
after multiple-comparison correction.
These results support our finding that outdated distractors
alone underestimate the difficulty posed by more diverse
distractors in our controlled setting.

\subsubsection{Significance of Post-Retrieval Failures}
\label{app:failure-significance}

\paragraph{Protocol.}
We test whether post-retrieval failures occur more frequently
than upstream failures in the reviewed cases.

For each model, we test
\[
H_0: p_{\mathrm{post}}=p_{\mathrm{upstream}},
\qquad
H_1: p_{\mathrm{post}}\neq p_{\mathrm{upstream}}.
\]
We apply two-sided exact McNemar tests, with Holm correction
across the two models at $\alpha=0.05$.
Models are tested separately to avoid treating their outputs
on shared questions as independent observations.
Pointwise 95\% confidence intervals are obtained from
200,000 paired bootstrap resamples stratified by dataset.

\paragraph{Results.}
Post-retrieval failures are significantly more prevalent than
upstream (pre-retrieval or retrieval) failures for both models,
evaluated with high reasoning effort and search. For GPT-5.4,
88.9\% of the reviewed failures involve a post-retrieval failure,
compared with 27.8\% involving an upstream failure
($p_{\mathrm{Holm}} = 0.0074$). For Gemini-3.1-Pro, the
corresponding rates are 100.0\% and 23.1\%
($p_{\mathrm{Holm}} = 3.81\times10^{-6}$). Because attribution
is multi-label, a case can count toward both categories. These
results support post-retrieval reasoning as the primary diagnosed
bottleneck.

\section{Details of \wild Experiments}
\label{app:wild_setup}

\subsection{Dataset Selection}
\label{app:subset-selection}

We focus on rapidly evolving knowledge using the source datasets' annotations.
For \textit{FreshQA}, we retain 49 questions labeled as both fast-changing and multi-hop.
For \textit{InfoDeepSeek}, we use 31 English questions marked as time-sensitive and freshness.
For \textit{SealQA}, we retain 64 fast-changing questions from the \texttt{seal\_hard} test split.
The resulting 144 questions are shared across all evaluated models.
We preserve the original reference answers, including subsequently identified annotation errors, which are examined separately during failure analysis.
Table~\ref{tab:dataset-time} summarizes the temporal reference years across the three subsets of \wild.
\begin{table}[t]
\centering
\footnotesize

\setlength{\tabcolsep}{6pt}
\begin{tabular}{@{}lrrr@{}}
\toprule
\textbf{Year} & \textbf{FreshQA} & \textbf{InfoDS} & \textbf{SealQA} \\
\midrule
2022        & 3  & 0  & 0 \\
2023        & 6  & 0  & $2^{\dagger}$ \\
2024        & 7  & 0  & 8 \\
2025        & 33 & 25 & 33 \\
2026        & 0  & 0  & 21 \\
Unspecified & 0  & 6  & 0 \\
\midrule
\textbf{Total} & \textbf{49} & \textbf{31} & \textbf{64} \\
\bottomrule
\end{tabular}

\smallskip
\parbox{\columnwidth}{\scriptsize
$\dagger$ Originally labeled ``before 2024''; grouped under
2023 as an upper-bound approximation.
}
\caption{Distribution of temporal reference years across the three subsets.}
\label{tab:dataset-time}
\end{table}

\subsection{Detailed information of the models and sub-datasets}
\label{app: model_list}
Table~\ref{tab:model-cutoffs} lists the evaluated models, their reported knowledge cutoffs, and official documentation links. 

\begin{table}
\centering
\footnotesize
\setlength{\tabcolsep}{4pt}
\renewcommand{\arraystretch}{1.05}

\resizebox{\columnwidth}{!}{%
\begin{tabular}{@{}lll@{}}
\toprule
\textbf{Model} & \textbf{Cutoff} & \textbf{Resource} \\
\midrule

\textsc{Claude-Sonnet-4.6}
& 2025-08
& \href{https://platform.claude.com/docs/en/models/sonnet-4-6/overview}{API docs} \\

\textsc{Claude-Opus-4.6}
& 2025-05
& \href{https://platform.claude.com/docs/en/models/opus-4-6/overview}{API docs} \\

\addlinespace[2pt]
\textsc{Gemini-3.1-Flash-Lite}
& 2025-01
& \href{https://ai.google.dev/gemini-api/docs/gemini-3}{API docs} \\

\textsc{Gemini-3.1-Pro}
& 2025-01
& \href{https://ai.google.dev/gemini-api/docs/gemini-3}{API docs} \\

\addlinespace[2pt]
\textsc{GPT-5.4-mini}
& 2025-08-31
& \href{https://developers.openai.com/api/docs/models/gpt-5.4-mini}{API docs} \\

\textsc{GPT-5.4}
& 2025-08-31
& \href{https://developers.openai.com/api/docs/models/gpt-5.4}{API docs} \\

\midrule

\textsc{Qwen3-8B}
& NR
& \href{https://huggingface.co/Qwen/Qwen3-8B}{Hugging Face} \\

\textsc{Qwen3-32B}
& NR
& \href{https://huggingface.co/Qwen/Qwen3-32B}{Hugging Face} \\

\textsc{Qwen3-235B}
& NR
& \href{https://huggingface.co/Qwen/Qwen3-235B-A22B}{Hugging Face} \\

\textsc{Qwen3.8-27B}
& NR
& \href{https://huggingface.co/Qwen/Qwen3.8-27B}{Hugging Face} \\

\textsc{GPT-oss-120B}
& 2024-06-01
& \href{https://huggingface.co/openai/gpt-oss-120b}{Hugging Face} \\

\textsc{GLM-4.7-Flash}
& NR
& \href{https://huggingface.co/zai-org/GLM-4.7-Flash}{Hugging Face} \\

\textsc{Gemma-4-31B}
& 2025-01$^\dagger$
& \href{https://huggingface.co/google/gemma-4-31B-it}{Hugging Face} \\

\bottomrule
\end{tabular}%
}
\caption{Reported cutoff dates and official model resources.
NR denotes a cutoff not explicitly reported in the cited documentation.
For Claude, we report the reliable knowledge cutoff;
$^\dagger$ denotes a pretraining-data cutoff.
Reasoning and search variants share the same base-model entry.}
\label{tab:model-cutoffs}
\end{table}

\subsection{Implement Details}
\label{app: details}
All open-source models are run locally on B200 GPUs using vLLM. The models can access the open web to retrieve necessary information through two tools: \texttt{web\_search} and \texttt{open\_url}. We use three types of tool backends: (i) native tools, such as the built-in web search tool available to GPT models in our experiments; (ii) externally available tools, such as Tavily\footnote{\url{https://www.tavily.com/}} and DuckDuckGo\footnote{\url{https://duckduckgo.com/}}; and (iii) model-mediated search. Since many native search tools do not expose APIs for direct external access, we use another model as a proxy to access these tools. For example, an open-source model can forward a search request to a GPT model, which invokes its native search tool on the requesting model's behalf. The proxy is instructed to return only a list of search results, discarding any generating an answer or reasoning trace, leaving the requesting model to interpret the results. This setup allows us to compare different search backends and better separate the contribution of the search tool from that of the LLM.

Unless otherwise specified, closed-source models use their native search tools, while open-source models use Tavily, which outperformed several alternative search tools (e.g. DuckDuckGo) in our preliminary experiments.

The Gemini family requires special handling because its native search tools do not expose the search results, preventing analysis with \tool{}. We therefore use OpenAI models as proxies to access their native search tools, following approach (iii). We also compare Gemini’s native search with the OpenAI-based proxy setup (Table~\ref{tab:gemini-search-comparison}), finding that the proxy achieves slightly higher accuracy on average.

\subsection{Costs Report}
\begin{table*}[t]
\centering
\footnotesize
\setlength{\tabcolsep}{5pt}
\begin{tabular}{@{}lrrrrr@{}}
\toprule
& \multicolumn{2}{c}{Tokens (k)}
& \multicolumn{3}{c}{Cost (USD)} \\
\cmidrule(lr){2-3}\cmidrule(l){4-6}
\textbf{Model}
& Input & Output
& Input & Output & Total \\
\midrule
\multicolumn{6}{c}{\textit{Closed-book LLMs}} \\

Claude-Sonnet-4.6
& 16.0 & 18.1 & 0.0 & 0.3 & 0.3 \\
Claude-Sonnet-4.6 \reason
& 16.0 & 283.2 & 0.0 & 4.2 & 4.3 \\
Claude-Opus-4.6
& 16.0 & 14.2 & 0.1 & 0.4 & 0.4 \\
Claude-Opus-4.6 \reason
& 16.0 & 236.6 & 0.1 & 5.9 & 6.0 \\

\addlinespace[2pt]
Gemini-3.1-Flash-Lite
& 15.6 & 0.5 & 0.0 & 0.0 & 0.0 \\
Gemini-3.1-Flash-Lite \reason
& 15.6 & 268.8 & 0.0 & 0.4 & 0.4 \\
Gemini-3.1-Pro
& 15.6 & 98.1 & 0.0 & 1.2 & 1.2 \\
Gemini-3.1-Pro \reason
& 15.6 & 407.4 & 0.0 & 4.9 & 4.9 \\

\addlinespace[2pt]
GPT-5.4-mini
& 14.9 & 3.1 & 0.0 & 0.0 & 0.0 \\
GPT-5.4-mini \reason
& 14.9 & 669.3 & 0.0 & 1.3 & 1.3 \\
GPT-5.4
& 14.9 & 2.5 & 0.0 & 0.0 & 0.0 \\
GPT-5.4 \reason
& 14.9 & 568.4 & 0.0 & 5.7 & 5.7 \\

\midrule
\multicolumn{6}{c}{\textit{Search-enabled LLMs}} \\

Claude-Sonnet-4.6
& 4,943.2 & 68.0 & 14.8 & 1.0 & 15.9 \\
Claude-Sonnet-4.6 \reason
& 5,868.1 & 395.8 & 17.6 & 5.9 & 23.5 \\
Claude-Opus-4.6
& 5,401.4 & 97.5 & 27.0 & 2.4 & 29.4 \\
Claude-Opus-4.6 \reason
& 6,067.9 & 442.6 & 30.3 & 11.1 & 41.4 \\

\addlinespace[2pt]
Gemini-3.1-Flash-Lite
& 15,242.1 & 54.3 & 3.8 & 0.1 & 3.9 \\
Gemini-3.1-Flash-Lite \reason
& 9,633.3 & 284.2 & 2.4 & 0.4 & 2.8 \\
Gemini-3.1-Pro
& 9,562.6 & 225.5 & 19.1 & 2.7 & 21.8 \\
Gemini-3.1-Pro \reason
& 30,630.3 & 780.0 & 61.3 & 9.4 & 70.6 \\

\addlinespace[2pt]
GPT-5.4-mini
& 1,548.4 & 16.7 & 0.4 & 0.0 & 0.4 \\
GPT-5.4-mini \reason
& 4,137.7 & 464.4 & 1.0 & 0.9 & 2.0 \\
GPT-5.4
& 1,479.1 & 14.8 & 1.8 & 0.1 & 2.0 \\
GPT-5.4 \reason
& 5,881.0 & 446.6 & 7.4 & 4.5 & 11.8 \\

\bottomrule
\end{tabular}
\caption{Token usage and estimated token costs on TREKBench-Wild,
aggregated over 144 questions per configuration.
Tokens are reported in thousands. All values are rounded to one decimal place;}
\label{tab:wild-token-cost}
\end{table*}
Table~\ref{tab:wild-token-cost} summarizes token usage and estimated token costs on TREKBench-Wild. Search and high reasoning effort generally increase costs. The total estimated token cost is \$250. Estimated search-call fees are \$50, while judging costs approximately \$1.

\subsection{Prompts}
\label{app:benchmark-prompts}

We provide the answer-generation and answer-grading prompts configured for
FreshQA, InfoDeepSeek, and SealQA. 

\subsubsection{FreshQA}
\label{app:benchmark-prompts-freshqa}

\paragraph{Answer generation.}

\begin{FrameworkPrompt}
Answer the question with only the final answer.

Answer as of {year}: the answer must reflect the state of the world in {year}. Interpret every relative time expression in the question — "current", "currently", "now", "today", "this year", "last year", "recent", "latest", "most recent" — as referring to {year} (use the end of {year} when a precise point is needed).

Question: {question}
Answer:
\end{FrameworkPrompt}

\paragraph{Answer grading.}
FreshQA first accepts deterministic equality with any normalized reference. Unresolved cases are sent to the following judge, which
returns \texttt{evaluation: correct} or \texttt{evaluation: incorrect}.
Multiple acceptable references are joined with \texttt{ | } in the
\texttt{\{answers\}} field.

\begin{FrameworkPrompt}
Judge whether the response answers the question correctly relative to at least
one acceptable reference answer.

Decision rules:
- Judge the primary/final answer, not its writing style or citation format.
- Mark correct when the response has the same meaning as any reference answer,
  uses a common alias or abbreviation, or logically entails a reference answer.
- Ignore articles, punctuation, harmless units, and trailing source citations.
- For numbers, require the exact value unless a reference itself is approximate
  or a range. When a reference is approximate or a range, accept a compatible
  approximation; reject a value that contradicts an explicit bound.
- Extra information makes the response incorrect only when it contradicts or
  materially changes the primary answer. Do not reject a correct answer merely
  because a citation is present.
- Use the supplied references as the evaluation target. Do not replace them
  with newer facts.

Return exactly one line:
evaluation: correct
or
evaluation: incorrect

Question: {question}
Effective year: {effective_year}
Acceptable reference answer(s): {answers}
Response: {response}
\end{FrameworkPrompt}

\subsubsection{InfoDeepSeek}
\label{app:benchmark-prompts-infodeepseek}

\paragraph{Answer generation.}
The InfoDeepSeek answer template contains the question without an additional
effective-year instruction. Temporal constraints stated in the question are
therefore preserved as written.

\begin{FrameworkPrompt}
Answer the question with only the final answer.

Question: {question}
Answer:
\end{FrameworkPrompt}

\paragraph{Answer grading: ordinary questions.}
InfoDeepSeek uses two judge templates.
The ordinary-question template below is selected unless the task metadata explicitly sets \texttt{false\_premise} to Boolean true. The routing uses the existing annotation, not a new LLM judgment about the question's premise. Both templates receive the question, reference answer(s), and candidate answer and return
\texttt{Yes} or \texttt{No}.

\begin{FrameworkPrompt}
Given a question and its ground-truth answer, determine whether
the candidate answer correctly answers the given question.
Pay attention to the following points:
1. The candidate answer must contain content that is
consistent with the ground-truth answer to be considered correct
(pay attention to the consistency of time, place, person, and
quantity), but more details can be provided.
2. If there are multiple contents/events/persons in the
ground-truth answer, the candidate answer must also contain
all the contents/events/persons to be considered correct.
3. The given question does not have a wrong premise, and
the relevant person/event must exist and be unique. If the
candidate answer proposes a wrong premise or cannot determine
whether the person/event exists, it is a wrong answer.
4. The given question has a clear answer and the given
ground-truth answer must be correct. If the candidate answer
does not answer the question correctly but proposes the need
to further query relevant information, it is a wrong answer.

Just answer Yes or No.

Given Question: {query}
Groundtrue Answer: {reference_answer}
Candidate Answer: {candidate_answer}
Does the candidate answer correctly answer the given question?
\end{FrameworkPrompt}

\paragraph{Answer grading: false-premise questions.}
For an annotated false-premise question, the specialized template requires the
candidate to identify or correct the same false premise as the reference.
If the reference also supplies an answer, the candidate must supply that answer
as well. This branch changes the grading prompt, not the generation prompt.

\begin{FrameworkPrompt}
Given a question and its ground-truth answer, determine whether the candidate answer correctly answers the given question.
Pay attention to the following points:
1. This question has a false premise, which has been pointed
out in the groundtruth answer. 
If the candidate answer does not point out or correct this false premise, it is incorrect.
2. If the false premise pointed out by the candidate answer is different from the groundtruth answer (time, place, event, person, meaning inconsistent), it is incorrect.
3. If the groundtruth answer still answers the question in addition to pointing out the false premise, the candidate answer should also answer the question. In this case, if the candidate answer does not answer the question or the meaning
of the answer content is inconsistent with the groundtruth answer (pay attention to the consistency of time, place, person, and quantity), it is incorrect.

Just answer Yes or No.

Given question: {query}
Groundtruth answer: {reference_answer}
Candidate answer: {candidate_answer}
Does the candidate answer correctly answer the given question?
\end{FrameworkPrompt}

\subsubsection{SealQA}
\label{app:benchmark-prompts-sealqa}

\paragraph{Answer generation.}
SealQA uses the same benchmark-level answer template as FreshQA. This is SealQA-Hard in the original paper, not LongSeal, which used a provided-document setting.

\begin{FrameworkPrompt}
Answer the question with only the final answer.

Answer as of {year}: the answer must reflect the state of the world in {year}. Interpret every relative time expression in the question — "current", "currently", "now", "today", "this year", "last year", "recent", "latest", "most recent" — as referring to {year} (use the end of {year} when a precise point is needed).

Question: {question}
Answer:
\end{FrameworkPrompt}

\paragraph{Answer grading.}
The following template includes the complete grading examples. The final
instruction requests a single letter: A for correct, B for incorrect, and C
for not attempted. Only A receives credit. Reference answers are inserted into
\texttt{\{target\}}, and the model output into
\texttt{\{predicted\_answer\}}.

\begin{FrameworkPrompt}
Your job is to look at a question, a gold target, and a predicted answer,
and then assign a grade of either ["CORRECT", "INCORRECT", "
NOT_ATTEMPTED"].
First, I will give examples of each grade, and then you will grade a new
example.
The following are examples of CORRECT predicted answers.
```
Question: What are the names of Barack Obama's children?
Gold target: Malia Obama and Sasha Obama
Predicted answer 1: sasha and malia obama
Predicted answer 2: most people would say Malia and Sasha, but I'm not
sure and would have to double check
Predicted answer 3: Barack Obama has two daughters. Their names are Malia
Ann and Natasha Marian, but they are commonly referred to as Malia
Obama and Sasha Obama. Malia was born on July 4, 1998, and Sasha was
born on June 10, 2001.
```
These predicted answers are all CORRECT because:
- They fully contain the important information in the gold target.
- They do not contain any information that contradicts the gold target.
- Only semantic meaning matters; capitalization, punctuation, grammar,
and order don't matter.
- Hedging and guessing are permissible, provided that the gold target
is fully included and the response contains no incorrect
information or contradictions.
The following are examples of INCORRECT predicted answers.
```
Question: What are the names of Barack Obama's children?
Gold target: Malia and Sasha
Predicted answer 1: Malia.
Predicted answer 2: Malia, Sasha, and Susan.
Predicted answer 3: Barack Obama does not have any children.
Predicted answer 4: I think it's either Malia and Sasha. Or it could be
Malia and Jackie. Or it could be Joey and Malia.
Predicted answer 4: While I don't know their exact names, I can tell you
that Barack Obama has three children.
Predicted answer 5: It's possible you may mean Betsy and Olivia. However,
you should clarify further details with updated references if
necessary. Is that the correct answer?
Predicted answer 6: It may be the case that Obama's child is named James.
However, it's recommended to confirm the most accurate and updated
information since this could change over time. This model may not
always reflect the most current information.
```
These predicted answers are all INCORRECT because:
- A factual statement in the answer contradicts the gold target.
Incorrect statements that have some hedging (e.g., "it is possible
that", "although i'm not sure, i think") are also considered
incorrect.
The following are examples of NOT_ATTEMPTED predicted answers.
```
Question: What are the names of Barack Obama's children?
Gold target: Malia and Sasha
Predicted answer 1: I don't know.
Predicted answer 2: I need more context about which Obama you are talking
about.
Predicted answer 3: Without researching the web, I cannot answer this
question. However, I can tell you that Barack Obama has two children.
Predicted answer 4: Barack Obama has two children. I know that one of
them is Malia, but I'm not sure about the other one.
```
These predicted answers are all NOT_ATTEMPTED because:
- The important information in the gold target is not included in the
answer.
- No statements in the answer contradict the gold target.
Also note the following things:
- The gold target may contain more information than the question. In such
cases, the predicted answer only needs to contain the information
that is in the question.
- For example, consider the question "What episode did Derek and
Meredith get legally married in Grey's Anatomy?" with gold target
"Season 7, Episode 20: White Wedding". Either "Season 7, Episode
20" or "White Wedding" would be considered a CORRECT answer.
- Do not punish predicted answers if they omit information that would be
clearly inferred from the question.
- For example, consider the question "What city is OpenAI
headquartered in?" and the gold target "San Francisco, California".
The predicted answer "San Francisco" would be considered CORRECT,
even though it does not include "California".
- Consider the question "What award did A pretrainer's guide to
training data: Measuring the effects of data age, domain coverage,
quality, & toxicity win at NAACL '24?", the gold target is "
Outstanding Paper Award". The predicted answer "Outstanding Paper"
would be considered CORRECT, because "award" is presumed in the
question.
- Do not give credit for an answer if it contains any internal
inconsistency.
- For example, consider the question: "How many NBA players have
scored 60 or more points in a regular season game since 2024?"
with the gold answer "8". A response is INCORRECT if it states "8
players" but lists 7 or 9, or if it initially says "8 players" but
later contradicts this by concluding 7 or 9.
Here is a new example. Simply reply with either CORRECT, INCORRECT, NOT
ATTEMPTED. Don't apologize or correct yourself if there was a mistake;
we are just trying to grade the answer.
```
Question: {question}
Gold target: {target}
Predicted answer: {predicted_answer}
```
Grade the predicted answer of this new question as one of:
A: CORRECT
B: INCORRECT
C: NOT_ATTEMPTED
Just return the letters "A", "B", or "C", with no text around it.
\end{FrameworkPrompt}

\subsection{Human Verification Guidelines}
\label{app:human-verification}

Review each case in the following order. Base decisions on
verifiable evidence

\begin{enumerate}

\item \textbf{Verify the reference answer and its temporal scope.}
First identify what the question asks, including the required
entities, conditions, and target time. Resolve relative expressions
such as ``current'' and ``last year'' using the benchmark's recorded
temporal convention, not the date of human review.
Check whether the reference answer is supported at that target time.
Distinguish when an event occurred or a claim was valid from when
a document was published, updated, or accessed. A newer page does
not necessarily describe a newer event.

\item \textbf{Verify the candidate answer and the judge's verdict.}
Evaluate the model's answer against the verified reference and
the question's requirements, then compare your decision with the
LLM judge's verdict. Apply the following rules:
\begin{itemize}[leftmargin=*]
    \item \textbf{Partial correctness is insufficient.}
    If the question requires multiple entities, facts, or conditions,
    all required components must be correct.
    \item \textbf{Valid alternatives are acceptable.}
    If several answers independently satisfy the question,
    any one is sufficient unless the question explicitly requests
    all of them. Do not confuse alternative acceptable answers
    with components of a required multi-item answer.
    \item Accept equivalent names, paraphrases, and harmless
    formatting differences. Reject incorrect entities, quantities,
    temporal qualifications, or contradictions affecting the answer.
    A correct answer mixed with incompatible guesses is not sufficient.
    \item For a verified false-premise question, the response must
    identify or correct the relevant false premise and provide any
    additional answer required by the reference.
\end{itemize}

\item \textbf{Determine whether the failure is system-independent.}
If the reference answer is invalid at the target time, record a
\emph{reference-answer/annotation error}. If the reference is valid
but the judge incorrectly rejects a correct response, record a
\emph{judge error}. Treat these as system-independent cases and
do not assign pipeline-stage labels to them in this review.
Proceed with system-related attribution only when the reference
is valid and the model's response is genuinely incorrect.

\item \textbf{Verify retrieval-side labels.}
Inspect the recorded search queries, the complete recorded retrieved
URL collection, and the evidence supporting the proposed matches,
not only the final-answer citations.
Confirm that a claimed equivalent document contains the relevant
answer-bearing facts under the question's entity, temporal, and
other constraints. Topical similarity alone is insufficient;
multiple retrieved documents may jointly provide the evidence.

If an exact gold-URL match or a valid answer-bearing equivalent
match is established, do not assign \emph{Pre-retrieval} or
\emph{Retrieval}. Otherwise, inspect the issued queries:
assign \emph{Pre-retrieval} when they fail to target the required
entity, time, or information need; assign \emph{Retrieval} when
they adequately target it, but the recorded results lack the evidence.
Treat query adequacy as a judgment about query formulation, not
proof that a search engine must return a particular page.

\item \textbf{Verify post-retrieval labels.}
Inspect the actual counterfactual inputs, answers, and grades.
Confirm that the probes use the evaluated model and the intended
temporal setting.

\begin{itemize}
    \item \textbf{Interpretation:} the gold-only answer is incorrect.
    Gold-plus-noise is not required after this failure.
    \item \textbf{Distraction:} gold-only is correct, but
    gold-plus-noise is incorrect.
    \item \textbf{Randomness/non-reproduction:} retrieval is adequate, and both probes are correct. This indicates that the original failure did not reproduce, which is not a statistically established cause.
\end{itemize}

Assess this branch independently of retrieval: an upstream failure does not rule out a downstream one.
\end{enumerate}

If your final judgment differs from the model’s attribution, identify the incorrect attribution and explain the specific reason for your disagreement.

\subsection{Human Verification and Agreement}
\label{sec:ver}
Two PhD students from China independently reviewed the verification results following the protocol in Appendix~\ref{app:human-verification}. Both reviewers were proficient in English, frequent internet users, and possessed the knowledge required for the task. Before reviewing, they carefully read the guidelines and received task-specific training. Each participant received compensation equal to US\$20 and provided informed consent for the use of their data. Compensation, working hours, and privacy protections complied with applicable local laws and regulations. The reviewers agreed on all items, suggesting that the judgments were grounded in clear, verifiable criteria rather than subjective preferences. Across 94 system-related error cases, human and model attributions showed strong agreement (Cohen's $\kappa = 0.930$).

\section{Details of \tool}
\label{app:failure-framework}

This appendix details the implementation of \tool, including its diagnostic
tests, attribution rules, human-verification interface, and prompts. 

\subsection{Inputs and Model Roles}
\label{app:framework-inputs}

Each input case contains a question, reference answer(s), ground-truth URLs, the
original model prediction, and the recorded search queries and returned URLs.
An effective year is included when available. The loader collects URLs from all
recorded search calls, rather than only final-answer citations, and removes exact
string duplicates while preserving their first occurrence. 

The implementation separates three model roles. A browsing diagnostic judge
extracts key evidence from gold pages and matches it against retrieved pages.
A non-browsing analysis judge evaluates query adequacy and grades counterfactual
answers. The \emph{model under evaluation} generates both counterfactual answers.
Reference answers are available to the diagnostic judges but are not included in the answering model's gold-only or gold-plus-noise prompts. 

\subsection{Retrieval-Side Tests}
\label{app:framework-retrieval}

\paragraph{URL identity matching.}
The first test compares gold URLs with retrieved URLs using a normalized identity.
Normalization ignores the URL scheme, removes fragments, lowercases the host,
removes a leading \texttt{www.}, and strips trailing slashes from the path. Path
case and query parameters are preserved. This is an identity heuristic rather
than a content comparison.

\paragraph{Answer-bearing evidence matching.}
Different URLs may contain the same decisive information. A browsing judge first
receives the question, reference answer(s), and gold URLs, and extracts a short
answer-bearing statement, preferably one or two sentences. If no URL identity
match exists and the retrieved collection is nonempty, a second call receives
this statement and the entire retrieved URL collection. It checks whether the
same facts occur in the supplied pages, allowing multiple pages to jointly supply
the evidence. Mere topical similarity is insufficient. The judge returns a
Boolean match, concise matching evidence, matching retrieved URLs, and an
explanation; it is not asked to produce another answer to the original question.

The prompt used for extracting short answer-bearing statements is listed below:

\begin{FrameworkPrompt}
You are extracting the key answer-bearing evidence for a temporal QA question.

Inputs are the question, the benchmark gold answer, and the supplied gold-document
URLs. Open and inspect only those gold URLs. Identify the shortest concrete evidence
that connects the question to the gold answer. Keep it to one or two short sentences;
do not create numbered requirements and do not evaluate the annotation.

Respond with ONLY a JSON object:

{
  "key_evidence": "brief answer-bearing fact or facts",
  "gold_evidence_urls": ["exact supplied gold URL", "..."]
}

---

Question: {question}
Effective year: {effective_year}
Gold answer(s): {gold_answers}

Gold-document URL(s):
{gold_urls}
\end{FrameworkPrompt}

Browsing diagnostic judge prompt is listed following:

\begin{FrameworkPrompt}
You are matching answer-bearing evidence between a gold document and a retrieved
web-document collection.

The short gold key evidence below was extracted in a separate pass from the
question, gold answer, and gold URLs. Use web access only to inspect URLs in the
supplied retrieved collection. Do not use outside URLs or general knowledge.

Look for the same key fact or facts in the retrieved collection. Multiple supplied
pages may jointly provide the evidence. `evidence_match` is true only when the
retrieved pages contain the gold key evidence, not merely when they mention the
same topic. Do not answer the original question. Return the concise matching
evidence and all matching retrieved URLs copied exactly from the input.

Respond with ONLY a JSON object:

{
  "evidence_match": true | false,
  "retrieved_key_evidence": "brief matching fact or facts, or empty",
  "equivalent_urls": ["exact supplied retrieved URL", "..."],
  "reason": "brief explanation"
}

---

Question: {question}
Effective year: {effective_year}
Gold answer(s): {gold_answers}
Gold key evidence: {gold_key_evidence}

Retrieved URL collection:
{retrieved_urls}
\end{FrameworkPrompt}

\paragraph{Query adequacy.}
When neither identity nor evidence matching succeeds, the analysis judge's query
assessment distinguishes pre-retrieval from retrieval failures. Adequate queries
should target the required entity, time, and fact; wrong entities or dates,
missing necessary multi-hop decomposition, or failure to target the answer are
grounds for inadequacy. A negative evidence match with adequate queries yields
\emph{Retrieval}; otherwise it yields \emph{Pre-retrieval}. This is a plausibility
assessment of the recorded queries, not a new search experiment establishing what
the queries would retrieve. Although the standard batch obtains this assessment
for every case, it affects attribution only when evidence matching fails.

The prompts used are listed below:

\begin{FrameworkPrompt}
You are grading two non-retrieval signals for a failed web-search QA case.
Produce factual signals only; a deterministic decision tree assigns the final
category. URL equivalence is checked separately by a browse-enabled audit.

You are given the gold answer and URLs, the gold-only probe result, the original
prediction, and the issued queries.

# Required judgements

1. `gold_only_correct`: whether the gold-only probe answer matches the gold
   answer, allowing trivial formatting or paraphrase differences.

2. `query_reasonable`: whether the queries reasonably target the required
   entity, time, and fact. Mark false for a wrong entity/time, missing multi-hop
   decomposition, or queries that never target the answer.

Respond with ONLY a JSON object:

{
  "gold_only_correct": true | false,
  "query_reasonable": true | false,
  "reason": "two or three concise, human-checkable sentences"
}

---

Question: {question}
Effective year: {effective_year}
Gold answer(s): {gold_answers}
Gold URL(s):
{gold_urls}

Gold-only answer: {gold_only_answer}
Gold page accessible: {page_accessible}
Original incorrect prediction: {prediction}

Search queries:
{search_queries}
\end{FrameworkPrompt}

\subsection{Post-Retrieval Counterfactuals}
\label{app:framework-counterfactuals}

\paragraph{Gold-only.}
The evaluated model receives the question, effective year, and gold URLs, with
instructions to answer from those pages. If a page is blocked, the prompt permits
recovering another representation of the same page, such as a cached copy or
search-result rendering. The model returns an answer, the contributing gold
URLs, a short explanation, and a page-accessibility flag. The analysis judge
compares the answer with the reference. An incorrect gold-only answer yields
\emph{Interpretation}, and no gold-plus-noise probe is run for that case.

It used the following prompt:

\begin{FrameworkPrompt}
You are running a controlled GOLD-ONLY probe for a failed web-search question.
Open and reason from only the supplied ground-truth page URL(s). Treat these
URLs as the authoritative context intended to answer the question. Do not use
unlisted sources or general web knowledge.

If a supplied URL is blocked or cannot be opened directly, recover an
accessible representation of that same page (for example a cached copy,
search-result rendering, mobile/alternate form, or mirror) and continue to
answer from the recovered page content.

Answer the question directly. Inspect every relevant supplied page and perform
any necessary entity selection, multi-hop reasoning, filtering, counting, date
calculation, or comparison. Return the best concrete answer derived from the
supplied URLs.

`page_accessible` reports whether the content of at least one supplied
ground-truth page was available directly or through such a representation.
`used_urls` must contain the original supplied URL(s) whose content contributed
to the answer.

Resolve relative time expressions against the effective year:
- "last year" means the effective year itself.
- "current", "currently", "this year", "most recent", "now", "latest", and
  "to date" mean as of the end of the effective year.

Respond with ONLY a JSON object:

{
  "rag_answer": "concrete best answer derived from the supplied pages",
  "page_accessible": true | false,
  "used_urls": ["exact supplied URL used", "..."],
  "rag_reasoning": "one or two concise sentences identifying the supporting fact"
}

---

Question: {question}
Effective year: {effective_year}

Ground-truth page URL(s):
{gold_urls}
\end{FrameworkPrompt}

\paragraph{Gold-plus-noise.}
This probe is run only when the gold-only answer is graded correct, regardless
of the retrieval-side diagnosis. The context contains the original retrieved
URL list followed by gold URLs. The prompt asks the model to answer from this
combined context, without identifying individual URLs as gold or noise. A failed
gold-plus-noise probe following a successful gold-only probe yields
\emph{Distraction}. 

It used the following prompt:

\begin{FrameworkPrompt}
You are running a controlled GOLD-PLUS-NOISE probe for a failed web-search
question. The URLs below are the complete allowed context. They may contain
relevant, irrelevant, conflicting, or outdated pages.

Open and reason from these URLs only. Do not use unlisted sources or general web
knowledge. Decide which evidence is relevant and current for the question, and
return the best concrete answer derived from the allowed context.

Resolve relative time expressions against the effective year:
- "last year" means the effective year itself.
- "current", "currently", "this year", "most recent", "now", "latest", and
  "to date" mean as of the end of the effective year.

Respond with ONLY a JSON object:

{
  "answer": "best concrete answer derived from the allowed context",
  "used_urls": ["exact allowed URL used", "..."],
  "reasoning": "one or two concise sentences identifying the decisive evidence"
}

---

Question: {question}
Effective year: {effective_year}

Allowed context URLs:
{context_urls}
\end{FrameworkPrompt}

\subsection{Multi-Label Attribution and Human Verification}
\label{app:framework-labels}

Final labels are assigned deterministically from the judge outputs, rather than
generated by a free-form classifier. The following pseudocode summarizes the
implemented rules. Here \texttt{E} denotes accepted evidence retrieval,
\texttt{Q} query adequacy, \texttt{G} gold-only correctness, and \texttt{N}
gold-plus-noise correctness. \texttt{unknown} denotes a missing or unparsed judgment.

\begin{FrameworkPrompt}
labels = {}
if not E:
    if Q is True:
        add Retrieval
    else:  # False or unknown
        add Pre-retrieval

if G is False:
    add Interpretation
elif G is True and noise_was_run and N is False:
    add Distraction

if labels is empty:
    add Randomness
\end{FrameworkPrompt}

A case may receive one retrieval-side label and one post-retrieval label, such as
Retrieval plus Distraction. Interpretation and Distraction are mutually exclusive
under the conditional probe schedule. With complete valid judgments, the
\emph{Randomness} category denotes an original failure that is not reproduced by
either probe and has no detected retrieval-side failure. However, the implementation
uses it as a residual category whenever no other rule fires, including some
missing-judgment cases. It is not a statistical estimate of stochastic failure.

\paragraph{Review artifacts.}
The JSONL output retains the question, answers, stage flags, query assessment,
all recorded retrieved URLs, matching URLs and evidence, counterfactual answers,
grades, and concise explanations. The review workbook separates the main checklist,
retrieval audit, and taxonomy, with fields for human verification and notes.
Because stage flags can overlap, per-category failure
counts are not mutually exclusive, and their percentages need not sum to 100\%.

\subsection{Execution and Reproducibility Details}
\label{app:framework-implementation}

The current driver defaults to GPT-5.4 with web search for gold-evidence extraction
and URL auditing, and GPT-5.4 without search for query assessment and answer grading.
These two configuration files do not explicitly set temperature or reasoning effort.
Documents are rendered with their index, URL, and text, with head truncation at 3,500 characters for non-gold pages and 50,000 for gold pages. Unavailable content is represented by a placeholder;

\begin{table*}[t]
\centering
\footnotesize                    
\setlength{\tabcolsep}{3pt} 
\begin{tabular}{l cc ccccc c}
\toprule
\multirow{2}{*}{\textbf{Benchmark}} 
& \multicolumn{2}{c}{\textbf{Temporal Dynamics}} 
& \multicolumn{5}{c}{\textbf{Real-World Challenges}} 
& \multirow{2}{*}{\textbf{\makecell{In-Depth \\ Analysis}}} \\
\cmidrule(lr){2-3} \cmidrule(lr){4-8}
& \makecell{Up-to-date \\ Knowledge} 
& \makecell{Evolving \\ Knowledge} 
& \makecell{Challenging\\Queries} 
& \makecell{Diverse\\Distractors} 
& \makecell{Challenging\\Distractors} 
& \makecell{Long\\Context} 
& \makecell{Frontier\\LRMs} 
& \\
\midrule
Magic Mushroom\\~\cite{zhang2025magic} & \xmark & \xmark & \xmark & \cmark & \xmark & \xmark & \xmark & \xmark \\
NoisyBench
\\~\cite{lee2026lost}     & \xmark & \xmark & \xmark & \cmark & \xmark & \xmark & \pmark & \xmark \\
HoH~\cite{ouyang2025hoh}            & \xmark & \cmark & \xmark & \xmark & \xmark & \xmark & \xmark & \xmark \\
LongSeal\\~\cite{pham2025sealqa}       & \pmark & \pmark & \cmark & \xmark & \xmark & \cmark & \cmark & \xmark \\
\midrule
\textbf{\dataset (Ours)} & \cmark & \cmark & \cmark & \cmark & \cmark & \cmark & \cmark & \cmark \\
\bottomrule
\end{tabular}
\caption{Comparison with existing benchmarks across temporal
dynamics, real-world challenges, and analysis depth. \dataset combines evolving-knowledge reasoning with diverse,
challenging distractors and in-depth failure analysis.
\cmark, \pmark, and \xmark indicate full, partial, and no
coverage, respectively.}
\label{tab:benchmark_comparison}
\end{table*}

\section{Dataset Construction Details}
\label{app:benchmark-construction}

\subsection{Questions and Evidence Collection}

We construct \dataset from the rapidly evolving knowledge subsets used in
\wild, retaining 49 FreshQA, 31 InfoDeepSeek, and 64 SealQA questions.
Each example contains the original question, reference answer(s), reference
year, gold documents, and a pool of distractors.
For SealQA, we reuse LongSeal's gold and outdated document snapshots as the initial evidence pool. Additional Relevance distractors are retrieved from real webpages, while Republication and Fabrication distractors are constructed from historical and gold evidence, respectively.
For FreshQA and InfoDeepSeek, we fetch the original reference URLs and cache their contents, retaining multiple gold documents when available.
Reference answers and existing human-review annotations are carried over rather than automatically updated during construction. We also distinguish the question's reference year from a document's date.
Relative expressions such as ``current'' and ``last year'' are interpreted
against the reference year, while explicit dates in the question are preserved.

\subsection{Distractor Generation}

\paragraph{Temporality.}
We target a pool of 20 historical documents per question.
For SealQA, we reuse LongSeal's supplied distractors and supplement shortages
from its larger document pool and additional web retrieval.
For FreshQA and InfoDeepSeek, a search-enabled LLM explores historical
query variants, after which we fetch candidate URLs and retain page text.

\paragraph{Relevance.}
We target 10 real webpages that are easy to confuse with evidence for the
original question. Given the question and reference answer, the search LLM
explores alternative constraints, including role, geographic scope, metric,
category, population, and ranking criterion.
Search begins with close neighbouring questions and broadens to related
queries when necessary. Candidate pages are fetched rather than represented
solely by generated search summaries.
A separate LLM verifier checks whether the passage supports a plausible competing answer of the appropriate type, while neither answering the original question nor supporting its reference answer.
We prioritize single-constraint mismatches but retain broader related candidates as a coverage fallback, recording which policy was used.

\paragraph{Trustworthiness: Republication.}
We target five republication distractors by assigning existing historical
documents newer displayed dates while retaining their substantive claims,
simulating old news republished to attract clicks~\cite{musi2022staying}.
When feasible, displayed dates fall after the gold-date anchor and within
the reference year. Otherwise, we use a late-reference-year fallback rather than omit the example.
If fewer than five source documents are available, a source may be reused;

\paragraph{Trustworthiness: Fabrication.}
We target five fabricated documents per question by minimally editing
question-relevant windows from gold documents, rather than unrelated
distractors~\cite{jawahar2022automatic}.
The generator proposes exact text replacements that introduce a plausible
incorrect answer through entity or value substitutions, together with any
dependent edits needed for consistency.
Unrelated wording and document structure are retained, and fabrication
does not itself refresh the publication date.
In the revised FreshQA/InfoDeepSeek pools, an additional step locates a
verbatim answer-bearing paragraph or table span before editing, excluding
reference titles, navigation, and recommendations. Generation and verification
also receive the reference year to avoid creating claims applicable only outside the question's time scope.

\subsection{Automated Quality Validation}

We retain document-like passages rather than reducing new distractors to
short generated summaries.
The target window length for Relevance and Fabrication is the median length
of up to ten nearby dated historical documents, bounded between 1,500 and
12,000 characters; when unavailable, the default is 6,000 characters.
Relevance passages are contiguous extracts from fetched pages, and
Fabrication windows preserve the surrounding gold text.
These targets reduce obvious length differences but do not impose identical lengths across types; 

Candidate discovery uses GPT-5.4-mini with web search, while semantic
verification and fabrication use GPT-5.4 with high reasoning effort.
Automated checks inspect missing fields, unusable pages, URL duplication,
date constraints, document counts, and verifier decisions.
Fabrication edits must match unique source spans, actually change the text,
declare an answer different from the reference, and retain at least 0.90 OpenAI text-embedding-3-small embedding similarity to the original window.
A separate verification call checks that the modified passage supports the declared incorrect answer rather than the reference and that its answer-relevant claims are internally consistent.

\subsection{Details of Dataset Setup}
\label{app:datasets_setup}
Gold documents are included in every condition and do not count toward the
distractor budget. We assemble four distraction conditions and a Gold-only
reference:
\begin{enumerate}
    \item \textbf{Outdated Only}: up to 20 historical documents plus gold evidence.
    \item \textbf{Mixed Natural Distractors}: up to 10 historical documents and
    10 Relevance distractors, plus gold evidence.
    \item \textbf{Source-Aware Adversarial}: a deterministic
    sample of up to 10 documents from the preceding mixed pool, plus up to
    five Republication and five Fabrication distractors, and gold evidence.
    \item \textbf{Source-Blind Adversarial}: the same documents in
    the same order as the preceding condition, with source URLs hidden.
    \item \textbf{Gold Only}: the gold documents without added distractors.
\end{enumerate}

Inspired by prior work on source credibility~\cite{ahdritz2025smel},
we assign synthetic documents to five suspicious fictitious domains:
\emph{totally-real-news.info}
\emph{definitely-not-fake.info}
\emph{trust-me-sources.info},
\emph{100-percent-accurate.info}
\emph{not-made-up.info}

These domains serve as source labels and are not used for web retrieval.
Each rendered document contains an index, date when known, and text.
The domain-cued condition additionally displays a source-URL field for both real and synthetic documents. 
The uncued condition omits this field;

Sampling and document shuffling are deterministic functions of the question ID and condition. We retain every question when a pool is incomplete and use the available documents instead of padding to the target count.
Consequently, 20 is a target distractor budget, not a fixed count for every example. 
Table~\ref{tab:construction-counts} reports the achieved average number of distractors per question; 

\begin{table}[t]
\centering
\footnotesize
\setlength{\tabcolsep}{2pt}

\resizebox{\columnwidth}{!}{%
\begin{tabular}{@{}lr rr rr rr rr@{}}
\toprule
& & \multicolumn{2}{c}{OO}
  & \multicolumn{2}{c}{MND}
  & \multicolumn{2}{c}{SAA}
  & \multicolumn{2}{c}{SBA} \\
\cmidrule(lr){3-4}
\cmidrule(lr){5-6}
\cmidrule(lr){7-8}
\cmidrule(l){9-10}
Subset & $N$ & $D$ & $L$ & $D$ & $L$
& $D$ & $L$ & $D$ & $L$ \\
\midrule
FreshQA & 49
& 16.27 & 40.6
& 16.39 & 41.0
& 17.88 & 43.5
& 17.88 & 43.0 \\
InfoDS & 31
& 15.61 & 47.2
& 14.48 & 43.4
& 17.58 & 51.7
& 17.58 & 51.1 \\
SealQA & 64
& 19.66 & 64.3
& 15.44 & 51.2
& 20.00 & 60.2
& 20.00 & 59.7 \\
\bottomrule
\end{tabular}%
}
\caption{Mean distractor counts ($D$, excluding gold) and context
lengths ($L$, in thousands of tokens, including gold) across conditions.
Lengths use \texttt{o200k\_base} and exclude questions and instructions.}
\label{tab:construction-counts}
\end{table}

Fetched pages, construction decisions, synthetic edit records, and assembled contexts are stored locally. Evaluation uses these frozen contexts without new web retrieval, so later webpage changes do not alter model inputs.

\subsection{Prompt}
\label{app:construction-prompts}

The following condensed instructions summarize the four construction
tasks. Republication is implemented programmatically rather than
through an LLM call.

\paragraph{Temporality.}
Historical-page discovery and verification use separate calls.
The subsequent date-only recovery step reuses cached pages without
rerunning the semantic verifier.

\noindent\textit{Candidate search.}
\begin{quote}
\footnotesize\ttfamily\raggedright
Use the web search tool to collect candidate URLs for historical evidence.\par
Question: \{question\}\par
Reference answer: \{gold\_answer\}\par
Find an older snapshot of the SAME evolving fact, strictly before \{cutoff\}.\par
We need \{needed\} more distinct URLs. Search the question's entities and relation,
then historical answer names and earlier years.\par
For this round, focus on pages BEFORE \{focus\_cutoff\}, not on the current answer.\par
Start with a historical version of the question plus before:\{focus\_cutoff\} \{exclude\_answer\}.\par
Next identify at least two plausible previous answers and search their names with
the central entity, the relevant relation and an earlier year. For example, for
an actor's latest film search dated reports about their previous film releases,
not a live filmography listing the current release. For a changing team roster,
search older roster announcements rather than the roster at the reference date.\par
Make at least THREE distinct search queries.

Strategy: \{strategy\}\par
Run several distinct queries and seek up to \{candidate\_k\} candidate real pages.\par
For current/latest/as-of-date qualifiers, seek an earlier version of the same fact.\par
Do not replace the central entity, role, event identity or ranking criterion.\par
The older answer should differ from the reference. Prefer dated news, official
historical records and contemporary profiles. Candidate discovery is not final
verification; include plausible historical pages even if a snippet is incomplete.\par
Do not invent URLs, page dates, or quotes. Exclude these URLs:\par
\{excluded\}\par
Return a JSON list of candidates with url, publication\_date (or empty if unknown),
title, supported\_answer (or empty), and evidence\_hint (or empty).
\end{quote}

The \texttt{\{strategy\}} field cycles through the following values:
\begin{quote}
\footnotesize\ttfamily\raggedright
1. Contemporary news reports stating the then-current winner, record holder, count, or ranking.\par
2. Official historical result pages and annual reports from earlier years.\par
3. Older year-specific rankings, lists, announcements, and retrospective reports.\par
4. Exact-phrase searches combining the central entities with prior years and prior values.\par
5. Historical versions and dated summaries from reputable reference or institutional sites.\par
6. Broader multilingual or regional reporting that still explicitly supports the old answer.
\end{quote}

\noindent\textit{Fetched-page verification.}
\begin{quote}
\footnotesize\ttfamily\raggedright
Read a retrieved historical page for a controlled evidence benchmark.\par
Question: \{question\}\par
Reference answer: \{gold\_answer\}\par
The page must be older than: \{cutoff\}\par
Candidate URL: \{url\}\par
Search-discovered date (unverified): \{publication\_date\}\par
Candidate historical answer (unverified): \{claimed\_answer\}\par
\smallskip
Actual fetched page text:\par
\{text\}\par
\smallskip
Does it explicitly support an earlier answer to the SAME evolving fact, different
from the reference? Interpret current/latest/as-of qualifiers at the earlier page
time, keeping the central entities, roles, scope and event identities unchanged.
A different event or unrelated fact is not an outdated answer. Do not invent an
old answer for a fixed historical fact. Extract the page's publication/update date
from visible page metadata or a dated news URL, not from arbitrary event dates,
copyright footer, or a search engine crawl time. An updated page supporting the
reference is not outdated just because its URL contains an old date.\par
Return ONLY JSON:\par
\{"answers\_exact\_question":true,
"claim\_supported":true,
"answer\_differs\_from\_gold":true,
"supported\_answer":"...",
"evidence":"verbatim supporting text",
"publication\_date":"YYYY-MM-DD or empty",
"date\_supported":true,
"date\_evidence":"verbatim date text or dated URL",
"reason":"brief explanation"\}
\end{quote}

\paragraph{Relevance.}
Candidate discovery and fetched-page verification are separate calls.

\noindent\textit{Candidate search.}
\begin{quote}
\footnotesize\ttfamily\raggedright
Find REAL web pages for a controlled requirement-mismatch benchmark.\par
Original question: \{q\}\par
Correct answer to the original question: \{gold\}\par
Answer cutoff year: \{year\}\par
Date policy for this search round: \{date\_policy\}\par
\smallskip
Prefer pages that directly answer a near-neighbour question obtained by changing
exactly ONE constraint in the original information need, such as role, entity
type, metric, scope, time, location, category, population, or ranking criterion.
If the search strategy is broader, pages may be weaker near neighbours, but they
must still be easy to mistake for answers to the original question when the
changed constraint is overlooked. Every page MUST explicitly support a concise
competing answer that differs from \{gold!r\} and has the same semantic answer type.
It must not directly answer the original question or support \{gold!r\} as that
question's answer.\par
\smallskip
Candidate-discovery strategy: \{search\_strategy\}\par
Excluded source URLs (never return these): \{excluded\}.\par
Search round: \{round\_no\}. Previously rejected URLs: \{rejected\}.\par
Return up to \{candidate\_k\} candidates. Return ONLY a JSON list with this schema:\par
[\{"url":"https://...",
"date":"YYYY-MM-DD",
"title":"...",
"near\_neighbor\_question":"...",
"mismatched\_constraint":"one concise constraint",
"supported\_answer":"the explicit competing answer",
"answer\_evidence":"a short verbatim phrase supporting that answer",
"relevance\_evidence":"a short phrase showing why it could be confused with the original need"\}].\par
If no valid page is found, return [].
\end{quote}

The \texttt{\{search\_strategy\}} field cycles through these alternatives:
\begin{quote}
\footnotesize\ttfamily\raggedright
near\_neighbor: Search explicit rewrites of the question that preserve its central
subject while changing one role, scope, metric, time, location, category,
population, or ranking constraint.\par
\smallskip
answer\_clusters: First derive two or three minimal near-neighbour questions, each
with a concrete answer different from the gold answer. Search for multiple
independent real pages that explicitly support those answers, so the competing
claims are corroborated.\par
\smallskip
constraint\_terms: Search the central entities together with alternative values
of one likely answer-determining constraint. Keep the original information need
otherwise fixed.\par
\smallskip
rank\_scope\_contrast: Target easily confused contrasts such as overall versus
subgroup, most versus latest, produced versus distributed, domestic versus
worldwide, or count versus percentage. Return only pages with an explicit answer
to the contrasted need.\par
\smallskip
keywords: Use broader keywords from the original question. A page may be a weaker
near neighbour, but it must state an explicit competing answer about the same
central subject that could be mistaken for the requested answer. Keyword overlap
alone is never sufficient.
\end{quote}

The \texttt{\{date\_policy\}} field requires dates strictly after the
gold-date anchor or its fallback, normally within the reference year.
The fallback search extends the upper bound through the following year;
per-question quotas are enforced by code.

\noindent\textit{Fetched-page verification.}
\begin{quote}
\footnotesize\ttfamily\raggedright
Classify a candidate distractor using only the supplied page text.\par
\smallskip
Original question: \{q\}\par
Correct answer: \{gold\}\par
Claimed near-neighbour question: \{near\_question\}\par
Claimed mismatch type: \{mismatch\_type\}\par
Claimed mismatched constraint: \{constraint\}\par
Claimed competing answer: \{supported\_answer\}\par
Claimed answer evidence: \{answer\_evidence\}\par
Claimed relevance evidence: \{relevance\_evidence\}\par
Candidate URL: \{url\}\par
Candidate date: \{date\}\par
Candidate title: \{title\}\par
\smallskip
Fetched page text:\par
\{text\}\par
\smallskip
Set relation to single\_constraint when the page answers a near-neighbour
information need that differs by exactly one answer-determining constraint;
related when it states a potentially distracting answer about the same central
subject but is not a strict single-constraint neighbour; otherwise invalid.
A page that directly answers the original question is invalid. A page that
supports the original gold answer is also invalid. The supplied passage must
explicitly support the claimed competing answer; that answer must differ from
the gold answer, have the same semantic answer type, and be plausible to return
if the changed constraint were overlooked.\par
\smallskip
Return ONLY JSON:\par
\{"relation":"single\_constraint|related|invalid",
"answers\_original\_question":false,
"supports\_gold\_answer":false,
"competing\_answer":"the answer explicitly supported by the passage",
"competing\_answer\_supported":true,
"competing\_answer\_same\_type":true,
"could\_mislead\_if\_constraint\_ignored":true,
"reason":"brief explanation"\}
\end{quote}

\paragraph{Republication.}

The procedure is shown in ~\Cref{alg:republication}. We assign replacement displayed dates and fictitious source URLs while retaining the source documents' substantive claims.  A source can be reused when fewer
than five are available.

The implementation additionally prioritizes
existing old-answer evidence and normalizes explicit publication metadata
and reader wrappers.  Date sampling is seeded
by the question ID.

\setlength{\intextsep}{4pt}
\begin{algorithm}[H]
\caption{Republication distractor construction}
\label{alg:republication}
\footnotesize
\begin{algorithmic}[1]
\Require Documents $D$, gold date $g$, reference year $y$
\Require Question ID $q$, revised-pipeline flag $r$
\Ensure Republished documents $R$

\State Remove empty and duplicate documents from $D$
\If{$D$ is empty}
  \State \Return $[]$
\EndIf

\State $e \gets \Call{YearEnd}{y}$
\State $s \gets$ ($g$ exists and $g<e$)
\If{$s$}
  \State $a \gets g$
\Else
  \State $a \gets e + 1\,\mathrm{day}$
\EndIf

\State $D \gets \Call{RankSources}{D,a,r}$
\State $S \gets$ first $\min(5,|D|)$ documents in $D$
\State Cycle through $S$ to fill five entries if needed

\If{$s$}
  \State $T \gets \Call{ForgedDates}{g,y,5,q}$
\Else
  \State $T \gets \Call{LateYearDates}{S,y,q}$
\EndIf

\State $R \gets []$
\For{$i=0,\ldots,4$}
  \State $d \gets \Call{Copy}{S_i}$
  \State $d.\mathrm{date} \gets T_i$
  \State $d.\mathrm{url} \gets \Call{SyntheticURL}{q,i}$
  \If{$r$}
    \State $d.\mathrm{text} \gets
      \Call{NormalizeMetadata}{d.\mathrm{text}}$
  \EndIf
  \State Save original URL, date, and reuse flag in $d$
  \State Append $d$ to $R$
\EndFor
\State \Return $R$
\end{algorithmic}
\end{algorithm}
Source ranking places dated pages before the anchor first, followed by
undated and remaining pages. In the revised implementation, old-answer
evidence is prioritized within these tiers.
Replacement dates normally fall between the day after the gold-date
anchor and the reference-year end; an unavailable interval triggers
the late-reference-year fallback.

\paragraph{Fabrication.}
The revised pipeline uses three calls: locating an answer-bearing span,
proposing minimal replacements, and verifying the modified passage.
Code applies the replacements and performs structural checks before
semantic verification.

\noindent\textit{Answer-span location.}
\begin{quote}
\footnotesize\ttfamily\raggedright
Locate the answer-bearing paragraph or table rows in the supplied article body.\par
Question: \{q\}\par
Correct answer: \{gold\}\par
Answer reference year: \{year\}\par
Interpret current/latest/last year relative to that year; preserve explicit dates
in the question. Locate the relevant claim or calculation inputs, not a newer
out-of-scope record. Return a verbatim contiguous span, including table headers
or neighbouring rows when needed. Ignore reference titles, navigation, and
recommendations. This is span location, not a judgment of the reference answer.\par
Article body:\par
\{body\}\par
\smallskip
Return ONLY JSON:
\{"answer\_span":"exact paragraph/table span, or empty if absent",
"kind":"paragraph|table"\}
\end{quote}

\noindent\textit{Minimal-edit generation.}
\begin{quote}
\footnotesize\ttfamily\raggedright
Create one synthetic hard-negative document by minimally editing the supplied
answer-bearing source window.\par
Question: \{q\}\par
Correct answer: \{gold\}\par
Answer reference year: \{year\}\par
Variant: \{variant\}\par
Attempt: \{attempt\}\par
Previously accepted competing answers (reuse is welcome): \{previous\}\par
Feedback from the previous rejected attempt: \{feedback\}\par
\smallskip
Ground-truth evidence:\par
\{evidence\}\par
\smallskip
Located answer-bearing paragraph/table (edit this claim first):\par
\{answer\_span\}\par
\smallskip
Source document window (copy is applied by code; do not return the full document):\par
\{source\_text\}\par
\smallskip
Identify the smallest set of exact text replacements that makes this document
explicitly answer the question with one plausible but incorrect answer.
Every old\_text must be a verbatim, uniquely occurring substring of the source
window. The wrong answer must differ from the correct answer and have the same
semantic type. Reusing a plausible competing answer is fine; distinct answers
across variants are not required. Vary source spans or formulations where
natural, not the answer solely for diversity. Interpret current/latest/last year
relative to \{year\}, while preserving explicit question dates. The wrong claim
must answer within that time scope, not merely become true in a later year.
Edit the located body paragraph or table, never a reference title, navigation,
or recommendation. Preserve all unrelated wording and document structure.
For counting or multi-hop questions, include every dependent replacement needed
for internal consistency. Do not edit the publication timestamp unless it is
itself the answer. Do not disclose that the result is synthetic or false.\par
\smallskip
Return ONLY JSON with this schema:\par
\{"wrong\_answer":"<incorrect answer explicitly supported after the edits>",
"edits":[\{"old\_text":"<exact unique source span>",
"new\_text":"<replacement span>"\}],
"changed\_claim":"<brief description of the exact factual edit>"\}
\end{quote}

\noindent\textit{Semantic verification.}
\begin{quote}
\footnotesize\ttfamily\raggedright
Audit a synthetic hard-negative document.\par
Question: \{q\}\par
Correct answer: \{gold\}\par
Answer reference year: \{year\}\par
Declared wrong answer: \{wrong\}\par
\smallskip
Modified passage:\par
\{modified\}\par
\smallskip
Judge only whether the modified document answers the exact question with the
declared wrong answer rather than the correct answer AS OF \{year\}, interpreting
relative dates from that year while respecting explicit question dates.
A claim that only answers in a later year does not qualify. The answer must
come from body prose or table data, not a reference title or recommended article.
Check whether all claims relevant to that answer are mutually consistent.
Ignore unrelated inconsistencies elsewhere in the long document.\par
\smallskip
Return ONLY JSON with a short reason:\par
\{"modified\_answers\_wrong\_only":true,
"modified\_is\_consistent":true,
"reason":"..."\}
\end{quote}

Rejected candidates receive the recorded error through
\texttt{\{feedback\}} on a subsequent attempt.
Candidates are accepted only when structural checks pass and both
verifier Boolean fields are true.

\newtcblisting{BaselinePrompt}[1]{
  title={#1},
  listing only,
  breakable,
  colback=gray!3,
  colframe=gray!40,
  fonttitle=\bfseries\small,
  coltitle=black,
  boxrule=0.4pt,
  arc=1mm,
  left=4pt,right=4pt,top=3pt,bottom=3pt,
  listing options={
    basicstyle=\ttfamily\scriptsize,
    breaklines=true,
    columns=fullflexible,
    keepspaces=true,
    showstringspaces=false
  }
}

\subsection{RAG Baseline Details in \dataset}
\label{app:rag-baselines}

All three baselines operate on the supplied documents without
additional retrieval or training. Let $q$ denote the question,
$y$ its effective year, $D$ the document collection, and $M$ the
evaluated model. Reranking and filtering use the shared answer
prompt below; Structured Conflict Resolution RAG uses its own contrastive answer prompt.

\paragraph{Reranking-Based RAG.}
We combine semantic relevance from \texttt{Qwen3-Reranker-8B}
with Gaussian temporal decay and pass the top 10 documents
to $M$. The cutoff $c$ is December 31 of year $y$.
Undated documents receive the mean temporal weight of dated
documents, or $1$ if none are dated.

\begin{algorithm}[H]
\caption{Joint Semantic--Temporal Reranking}
\label{alg:ctrl-reranking}
\footnotesize
\begin{algorithmic}[1]
\Require Question $q$, year $y$, documents $D$, model $M$
\For{each document $d_i \in D$}
    \State Split $d_i$ into 256-word chunks with 32-word overlap
    \State $s_i \gets \max_{x \in \mathrm{Chunks}(d_i)}
        \mathrm{Reranker}(q,y,x)$
    \State $\Delta_i \gets c-\mathrm{date}(d_i)$, in days
    \State $w_i \gets 2^{-(\Delta_i/365)^2}$ if $\Delta_i \geq 0$,
        otherwise $0$
\EndFor
\State Impute missing $w_i$ using the rule above
\State $D' \gets$ top $\min(10,|D|)$ documents by $s_iw_i$
\State \Return $M(\mathrm{AnswerPrompt}(q,y,D'))$
\end{algorithmic}
\end{algorithm}

\begin{BaselinePrompt}{Reranking Query}
Find evidence that satisfies every requirement of the question.
Question: {question}
Answer time frame: {effective_year}
\end{BaselinePrompt}

\paragraph{ Filtering-Based RAG.}
The model selects documents satisfying the question's requirements,
removing irrelevant, superseded, or untrustworthy information.
A second call answers using the retained documents, with no
fixed limit on their number.

\begin{algorithm}[H]
\caption{LLM Context Filtering}
\label{alg:ctrl-filtering}
\footnotesize
\begin{algorithmic}[1]
\Require Question $q$, year $y$, documents $D$, model $M$
\State $r \gets M(\mathrm{FilterPrompt}(q,y,D))$
\State $I \gets$ valid, unique document IDs parsed from $r$
\If{$I$ is empty}
    \State $D' \gets D$
\Else
    \State $D' \gets$ documents indexed by $I$
\EndIf
\State \Return $M(\mathrm{AnswerPrompt}(q,y,D'))$
\end{algorithmic}
\end{algorithm}

\begin{BaselinePrompt}{Context Filtering Prompt}
Filter the documents before another model answers the question as of
{effective_year}. Keep documents containing information that directly satisfies
the complete information need. Remove documents that are irrelevant, fail any
requirement in the question, or contain untrustworthy information. If multiple
documents contain information that satisfies the same need, retain the latest
information that is valid for the requested time and remove superseded versions.
There is no required or maximum number of documents to keep. Do not answer the
question.

Return only JSON with all one-based document IDs that should be retained:
{"keep_document_ids": [2, 5, 9]}

Question: {question}

{docs_block}
\end{BaselinePrompt}

\paragraph{Structured-Evidence RAG}
Our lightweight, two-stage adaptation first extracts
document-level evidence and candidate answers, then compares
these records to resolve conflicting claims.

\begin{algorithm}[H]
\caption{Structured Conflict Resolution}
\label{alg:ctrl-crag}
\footnotesize
\begin{algorithmic}[1]
\Require Question $q$, year $y$, documents $D$, model $M$
\State $E \gets M(\mathrm{ExtractPrompt}(q,y,D))$
\Statex \hspace{\algorithmicindent}
    Each record contains a document ID, key evidence,
\Statex \hspace{\algorithmicindent}
    candidate answer, relation to $q$, and brief reason.
\State $(b,a) \gets M(\mathrm{ResolvePrompt}(q,y,E))$
\State \Return answer $a$, with brief comparison $b$ recorded
\end{algorithmic}
\end{algorithm}

\begin{BaselinePrompt}{C-RAG: Evidence Extraction}
For every document, extract the passage most useful for answering the
question and identify the answer it supports. Briefly state whether it
directly answers the question, provides competing evidence, or is irrelevant.
Do not answer the question.

Return only a compact JSON array with these fields: document_id,
key_evidence, candidate_answer, relation, and reason.

Question: {question}
Time frame: {effective_year}

{docs_block}
\end{BaselinePrompt}

\begin{BaselinePrompt}{C-RAG: Conflict Resolution}
Using the extracted evidence, compare the candidate answers and determine
which one best satisfies the question as of {effective_year}. Briefly explain
why it is better supported than the competing answers.

Return exactly:

#Contrast: <brief comparison>
#Answer: <short final answer>

Question: {question}

Extracted evidence:
{extracted_evidence}
\end{BaselinePrompt}

\paragraph{Shared Answer Prompt.}
Direct generation uses this prompt with all supplied documents;
reranking and filtering use it with their selected documents.

\begin{BaselinePrompt}{Shared Answer Generation}
Answer the question using ONLY the documents provided below. Give only the final answer, with no explanation.

Time frame: answer as of {effective_year}. Interpret every relative time expression in the question ("current", "currently", "now", "today", "this year", "last year", "recent", "latest", "most recent") as referring to {effective_year}.

The documents are listed in no particular order and each is labelled with its date when known. Because this fact changes over time, several documents will report different, conflicting values for it. Your task is to find the value that is both the most recent and relevant as of {effective_year}:
- Among the conflicting values, identify the latest one that still holds as of {effective_year}, using the document dates to order them.
- Ignore values that a later, relevant document has since superseded, and ignore documents that do not actually address the question.

=== DOCUMENTS ===
{docs_block}
=== END OF DOCUMENTS ===

Question: {question}
Answer:
\end{BaselinePrompt}

\section{Comparison with existing distraction benchmarks}

\label{app:more_related_work}

As summarized in Table~\ref{tab:benchmark_comparison},
our work advances existing evaluations by combining temporal
dynamics, diverse real-world challenges, and in-depth failure
analysis for reasoning over rapidly evolving knowledge. Existing benchmarks emphasize either generic noise without
explicit temporal control~\cite{zhang2025magic,lee2026lost}
or outdated evidence that can enable recency
shortcuts~\cite{ouyang2025hoh,pham2025sealqa}.

We exclude datasets whose data are not publicly available, such as RecencyQA~\citep{piryani2026often}.

\section{Check List}
\subsection{Risk}
We identify no significant ethical or safety risks associated with our approach, as it operates on publicly available data. We only used publicly available datasets that do not contain personally identifiable information or offensive content, as verified by the dataset providers.

TrekBench-Ctrl contains synthetic documents that either state false claims about real entities (Fabrication) or present outdated content under a newer date (Republication). Outside the benchmark, such documents could be mistaken for genuine reports. We limit this risk in four ways. First, every synthetic document is marked as synthetic in the released data and stored together with its source document and the exact edits applied, so each false claim can be traced and reverted. Second, synthetic documents are attributed only to fictitious domains that are implausible by design (e.g., totally-real-news.info), never to real outlets. Third, fabrications are minimal entity or value substitutions tied to a single benchmark question rather than new narratives, which makes them of little use as standalone misinformation. Fourth, we release the synthetic documents only as part of the frozen benchmark contexts, under terms that restrict their use to research and evaluation and do not permit redistributing them separately.

\subsection{The License For Artifacts}

All models and datasets used in this work comply with their respective open-source or research licenses. We ensure that all artifacts are used strictly within the permitted scope of their terms. The Code we released will be under a permissive open-source license, enabling reproducibility and reuse.

\subsection{Ai Assistants}
We used AI assistants (ChatGPT) solely for textual and grammatical refinement, without influencing the core content or experimental results.

\end{document}